\documentclass[a4paper,fleqn]{cas-dc}
\usepackage{bbding}
\usepackage[numbers]{natbib}
\usepackage{amsmath}
\usepackage{amssymb}
\usepackage{multirow}
\usepackage{makecell}
\usepackage{wrapfig}
\usepackage{flushend}
\usepackage{setspace}
\usepackage{xcolor}

\def\tsc#1{\csdef{#1}{\textsc{\lowercase{#1}}\xspace}}
\tsc{WGM}
\tsc{QE}
\tsc{EP}
\tsc{PMS}
\tsc{BEC}
\tsc{DE}

\begin{document}
\let\WriteBookmarks\relax
\def\floatpagepagefraction{1}
\def\textpagefraction{.001}

\shorttitle{Multimodal Graph Representation Learning for Robust Surgical Workflow Recognition with Adversarial Feature Disentanglement}

\shortauthors{L. Bai et~al.}

\title [mode = title]{Multimodal Graph Representation Learning for Robust Surgical Workflow Recognition with Adversarial Feature Disentanglement}                      
\tnotemark[1]

\tnotetext[1]{This document is the results of the research projects by Hong Kong Research Grants Council (RGC) Research Impact Fund (RIF) R4020-22, Collaborative Research Fund (CRF) C4026-21GF, General Research Fund (GRF 14203323, GRF 14216022, and GRF 14211420),  NSFC/RGC Joint Research Scheme N\_CUHK420/22; Shenzhen-Hong Kong-Macau Technology Research Programme (Type C) STIC Grant 202108233000303; Guangdong Basic and Applied Basic Research Foundation \#2021B1515120035.}

\author[1,2]{Long Bai}[orcid=0000-0002-9762-6821]
\fnmark[1]

\ead{b.long@link.cuhk.edu.hk}
\credit{Conceptualization, Methodology, Formal analysis, Investigation, Resources, Validation, Writing — original draft preparation, Project administration}

\affiliation[1]{organization={Department of Electronic Engineering, The Chinese University of Hong Kong},
    city={Hong Kong},
    country={China}}

\affiliation[2]{organization={Chair for Computer Aided Medical Procedures, Technical University of Munich},
    city={Munich},
    country={Germany}}

\author[1,3]{Boyi Ma}[orcid=0009-0001-6913-6171]
\fnmark[1]
\ead{boyima.ma@mail.utoronto.ca}
\credit{Methodology, Formal analysis, Investigation, Validation, Writing - original draft preparation, Visualization}

\affiliation[3]{organization={Department of Biomedical Engineering, University of Toronto},
    city={Toronto},
    country={Canada}}

\author[4]{Ruohan Wang}[orcid=0009-0005-2146-2292]
\ead{ruohan_wang@brown.edu}
\credit{Conceptualization, Methodology, Formal analysis, Investigation, Validation}
\affiliation[4]{organization={Center for Computational and Molecular Biology, Brown University},
    city={Providence},
    country={United States}}

\author[1]{Guankun Wang}[orcid=0000-0003-2440-4950]
\ead{gkwang@link.cuhk.edu.hk}

\credit{Methodology, Investigation, Writing — review \& editing, Data curation}

\author[1]{Beilei Cui}[orcid=0009-0009-7900-8032]
\ead{beileicui@link.cuhk.edu.hk}

\credit{Methodology, Investigation, Writing — review \& editing, Data curation}

\author[2]{Zhongliang Jiang}[orcid=0000-0001-7461-2200]
\ead{zl.jiang@tum.de}

\credit{Methodology, Investigation, Writing — review \& editing}

\author[5]{Mobarakol Islam}[orcid=0000-0002-7162-2822]
\ead{mobarakol.islam@ucl.ac.uk}

\credit{Conceptualization, Methodology, Investigation, Writing — review \& editing}
\affiliation[5]{organization={UCL Hawkes Institute, University College London},
    city={London},
    country={UK}}

\author[5,6]{Zhe Min}[orcid=0000-0002-8903-1561]
\ead{minzhe@sdu.edu.cn}

\credit{Conceptualization, Methodology, Investigation, Writing — review \& editing}

\affiliation[6]{organization={School of Control Science and Engineering, Shandong University},
    city={Jinan},
    country={China}}

\author[1]{Jiewen Lai}[orcid=0000-0002-2676-7387]
\ead{jiewen.lai@cuhk.edu.hk}

\credit{Conceptualization, Investigation, Writing — review \& editing, Supervision}

\author[2]{Nassir Navab}[orcid=0000-0002-6032-5611]
\ead{nassir.navab@tum.de}

\credit{Methodology, Investigation, Writing — review \& editing, Supervision}

\author
[1]{Hongliang Ren}[orcid=0000-0002-6488-1551]
\cormark[1]

\ead{hlren@ee.cuhk.edu.hk}
\credit{Conceptualization, Resources, Writing — review and editing, Supervision, Project administration, Funding acquisition}

\cortext[cor1]{Corresponding author: H. Ren.}

\fntext[fn1]{Co-first authors: L. Bai and B. Ma.}

\begin{abstract}
Surgical workflow recognition is vital for automating tasks, supporting decision-making, and training novice surgeons, ultimately improving patient safety and standardizing procedures. However, data corruption can lead to performance degradation due to issues like occlusion from bleeding or smoke in surgical scenes and problems with data storage and transmission. Therefore, a robust workflow recognition model is urgently needed. In this case, we explore a robust graph-based multimodal approach to integrating vision and kinematic data to enhance accuracy and reliability. Vision data captures dynamic surgical scenes, while kinematic data provides precise movement information, overcoming limitations of visual recognition under adverse conditions. We propose a multimodal Graph Representation network with Adversarial feature Disentanglement (GRAD) for robust surgical workflow recognition in challenging scenarios with domain shifts or corrupted data. Specifically, we introduce a Multimodal Disentanglement Graph Network (MDGNet) that captures fine-grained visual information while explicitly modeling the complex relationships between vision and kinematic embeddings through graph-based message modeling. To align feature spaces across modalities, we propose a Vision-Kinematic Adversarial (VKA) framework that leverages adversarial training to reduce modality gaps and improve feature consistency. Furthermore, we design a Contextual Calibrated Decoder, incorporating temporal and contextual priors to enhance robustness against domain shifts and corrupted data. Extensive comparative and ablation experiments demonstrate the effectiveness of our model and proposed modules. Specifically, we achieved an accuracy of 86.87\% and 92.38\% on two public datasets, respectively. Moreover, our robustness experiments show that our method effectively handles data corruption during storage and transmission, exhibiting excellent stability and robustness. Our approach aims to advance automated surgical workflow recognition, addressing the complexities and dynamism inherent in surgical procedures.
\end{abstract}


\maketitle

\section{Introduction}

The advent of robot-assisted minimally invasive surgery (RMIS) has marked a transformative era in modern medicine, offering excellent precision, stability, and control that significantly exceeds the capabilities of the human hand~\cite{cui2024surgical,han2022systematic,wang2024copesd}. Robotic systems, such as the da Vinci Surgical System, have been widely adopted across various surgical disciplines~\cite{pugin2011history,wang2024magnetic}, demonstrating significant benefits like minimally invasive procedures, reduced patient recovery times, and improved surgical outcomes~\cite{chen2024lightdiff,wang2024surgical,wang2024video}. These sophisticated robots act as extensions of the surgeon's expertise, enabling complex operations to be performed through small incisions with enhanced visualization and dexterity~\cite{gao2023transendoscopic,li2023three}. As the adoption of robotic-assisted surgery continues to grow, the demand for intelligent systems that can recognize and understand surgical workflows becomes increasingly significant~\cite{bai2023surgical,chen2024surgsora,moglia2021systematic}.

Artificial Intelligence (AI)-assisted workflow recognition in robotic surgery serves multiple essential purposes~\cite{bai2023cat,jin2017sv,jin2021temporal,psychogyios2023sar}. It is the key to automating routine tasks, providing decision-making support, and enabling context-aware assistance during operations~\cite{bai2024ossar,wang2023domain}. Furthermore, understanding surgical workflows is crucial for training junior surgeons, as it enables the objective assessment of technical skills and procedural knowledge~\cite{long2021relational}. By analyzing the sequence and structure of surgical tasks, workflow recognition systems can provide real-time feedback, facilitate error correction, and contribute to the standardization of surgical procedures, ultimately leading to improved patient safety and better training methodologies~\cite{cao2023intelligent}.

\begin{figure}[t]
    \centering
    \includegraphics[width=0.9\columnwidth]{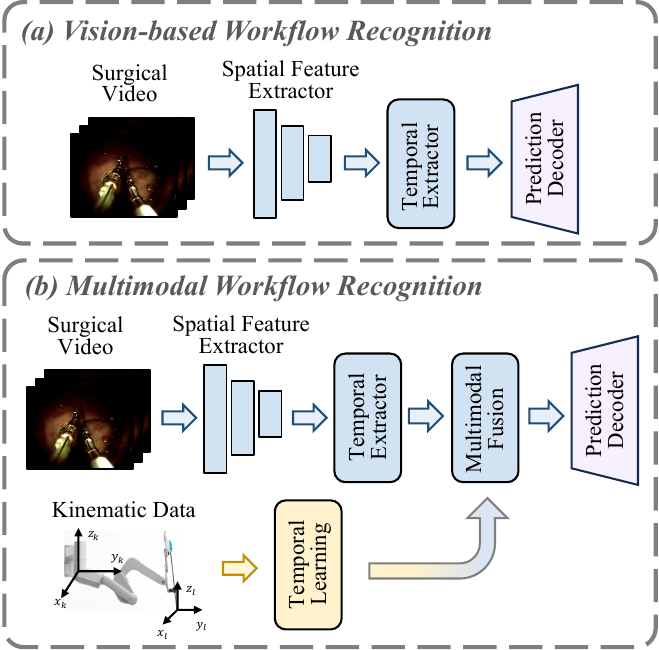}
    \caption{Comparison between the vision-based workflow recognition framework and the multimodal workflow recognition framework.}
    \label{fig:intro}
\end{figure}

Vision-based surgical workflow recognition has significantly advanced~\cite{twinanda2016endonet}. Vision data, derived from high-resolution cameras, captures the dynamic surgical scene, allowing for the visual identification of instruments, tissue interactions, and operative gestures~\cite{jin2022trans, yuan2023learning}. These methods show promise in accurately identifying surgical phases and tool usage. However, they face challenges in complex scenarios, such as occlusions or variable lighting. Similarly, kinematic-based recognition systems focus on the mechanical aspects of surgery, analyzing movement patterns and instrument trajectories to deduce ongoing tasks~\cite{forestier2018surgical}. While these systems are less affected by visual interference, they often lack the context that visual cues provide, which is crucial for a comprehensive understanding of the surgical workflow. Kinematic systems might also misinterpret complex instrument interactions. In this case, despite advancements in single-modality systems, their isolated application is often inadequate for capturing the full spectrum of surgical scenes. This necessitates exploring multimodal workflow recognition approaches that take advantage of both visual and kinematic data for enhanced performance and reliability.

Currently, the fusion of vision and kinematic data presents a compelling approach that promises enhanced accuracy and reliability~\cite{long2021relational}. Visual data effectively provides the context and semantic information needed for workflow analysis. Meanwhile, kinematic data, which includes the precise movements and positions of robotic arms and instrument tips, is immune to visual obstructions and offers additional information not always discernible through vision alone. These two data modalities complement each other, with vision providing context and kinematic data offering unambiguous motion patterns~\cite{huaulme2021micro,long2021relational}. Therefore, as presented in Figure~\ref{fig:intro}, a multimodal system shall enable a more complete and detailed interpretation of surgical activities, leading to a robust workflow recognition system that can adapt to the complex and variable nature of surgical environments.

However, current vision-kinematics methods primarily concentrate on fusing different modalities without adequately exploring the rich features within each modality. This limitation can impede the model's ability to comprehend complex surgical scenarios, as it may not capture the unique higher-level semantic information inherent to each data type. Optimizing each modality separately can more effectively uncover and leverage the intrinsic characteristics of each data type, thereby enhancing overall performance.
Moreover, in surgical and endoscopic environments, data quality may degrade due to camera movement, occlusions, or noise and quality loss during data transmission. Therefore, robust automatic recognition of surgical workflows is crucial to deal with noise and disturbances typical in surgical settings. Current approaches might not sufficiently address the need for robustness, potentially leading to unstable performance in real-world operating environments. Hence, our method needs to be robust enough to handle explicit data corruption issues, ensuring clinical-level accuracy.
These identified weaknesses highlight the necessity for improvements in multimodal data processing and robustness in surgical workflow recognition systems. We will focus on strategies that enhance robustness through adversarial discrimination and graph calibration loss. Additionally, we will further strengthen the feature representation of visual nodes through disentanglement to better handle the complexity and dynamism of surgical procedures. Such strategies would contribute to advancing the accuracy and stability of automated surgical workflow recognition.

To tackle the above challenges, we propose a multimodal \textbf{G}raph \textbf{R}epresentation network with \textbf{A}dversarial feature \textbf{D}isentanglement (\textbf{GRAD}) to perform the surgical workflow recognition with vision and kinematics data.
We first developed a Multimodal Disentanglement Graph Network (MDGNet) to explore and model the intra-modal and inter-modal feature connections between vision and kinematic modalities. Initially, we explore the unique features of each modality to enhance internal feature learning. For visual data, we delve deeply into disentangling features across three domains: spatial, wavelet, and Fourier, and propose the disentanglement graph learning strategy. This multi-domain approach allows us to disentangle complex patterns and representations that may not be apparent in the raw data. To capture the dynamic aspects of visual data, we employ Temporal Convolutional Networks (TCNs), specifically designed to extract time-related characteristics from sequential data. For kinematics data regarding object motion, we use both TCNs and Long Short-Term Memory (LSTM) networks to extract temporal movement features. By fully decoupling intra-modal information, we can effectively distinguish and utilize the independence and complementarity of each data modality. Subsequently, the attention-based graph network models the kinematic information and visual data as nodes to learn relational representations, while the attention mechanism strengthens interdependencies in complex data.

Moreover, to impose more effective constraints on the multimodal feature space, and to align the distributions of different modalities, we propose a Vision-Kinematic Adversarial (VKA) learning strategy. By optimizing a discriminative joint feature space, our strategy brings the multimodal feature distributions closer together, thereby reducing inter-modal discrepancies.
Finally, we propose our contextual calibrated decoder to maintain high robustness against data corruption. Graph networks may have low confidence due to the shallow layers. We integrate graph embeddings with the original multimodal feature embeddings and use a regularized calibration optimization function to address prediction confidence issues, significantly improving the adaptiveness and stability of our model when dealing with corrupted data (e.g., data with noise, blur, occlusion, or digital corruptions).
The main contributions of this work are as follows:
\begin{itemize}
    \item[--] We present the GRAD framework, a robust graph representation learning framework for surgical workflow recognition using visual and kinematic data. Our architecture leverages the MDGNet to establish interactions between different modalities based on in-depth information exploration within each modality, providing an effective solution for surgical workflow analysis.
    \item[--] We further incorporate Vision-Kinematic Adversarial (VKA) learning, which forces the model to align multimodal feature distribution during the pre-fusion stage. Additionally, our contextual calibrated decoder further strengthens the output confidence and enhances robustness.
    \item[--] We validate our proposed GRAD framework on two publicly available robotic surgery datasets. Our approach achieves superior performance compared to both single-modality and multimodal methods. Our ablation studies further demonstrate the effectiveness of the proposed components. Moreover, in robustness tests, our method shows more stable performance and slower performance degradation, proving its real-world clinical efficacy.
\end{itemize}

\section{Related Work}

\subsection{Multimodal Graph Learning}
Over the past several years, graph representation learning has gained significant attention in many different domains~\cite{nong2022adaptive,chen2023multimodal,wang2021dualgnn,mai2023multimodal,zhu2021dyadic,zhu2024graph}. The relationships and dependencies between different modalities can be better represented graphically, therefore, graph representation learning has been verified as an effective way to implement multi-feature fusion. In recent studies, an increasing number of multimodal fusion tasks demonstrated great performance with graph representation learning in the medical domain. Zeng et al.~\cite{zeng2023multimodal} utilized a multilayer graph convolutional network (GCN) to fuse multimodal information, effectively capturing long-range contextual dependencies in the text while filtering out visual noise unrelated to the textual content. Zhu et al.~\cite{zhu2023brain} introduced a feature fusion mechanism that grouped pixels with similar characteristics into nodes and transformed spatial domain input features into the graph domain. Hou et al.~\cite{RN923} developed a hybrid GCN that combines the strengths of traditional GCNs and hypergraph convolutional networks (HCNs). This model leverages node message passing to enable both intra-modal and inter-modal interactions within multimodal graphs. Yue et al.~\cite{yue2021multimodal} improved fusion effectiveness by employing a GCN to integrate diverse features from multiple data modalities, utilizing graph topology to propagate messages and represent node features along with their interrelations through the graph structure.
Qu et al.~\cite{qu2021ensemble} considered subjects' relationships within and between modalities by learning a graph embedding from multimodal data. Li et al.~\cite{li2024multimodal} suggested a 3D Haar semi-tight graph learning method to integrate data from various modalities, e.g., vision, text, audio, and statistical information. Meng et al.~\cite{meng2024masked} utilized a masked graph network to integrate various modalities for emotion recognition and prediction.

Currently, initial works use graph learning to fuse and model the multimodal perception information in surgical scenarios. However, existing work mainly employs simple multimodal modeling using graph-based methods without deeply exploring the performance of different modalities in surgery or researching more robust graph solutions. Therefore, we aim to develop effective and robust graph representation learning methods to further enhance the real-world application potential of surgical workflow recognition.

\subsection{Surgical Workflow Recognition}
Automatic surgical workflow recognition is vital in modern intelligent operating rooms as it provides context awareness for computer-assisted systems~\cite{ayobi2024pixel,liu2018deep,liu2025lovit,liu2023skit,yuan2021surgical,yuan2022anticipation,yuan2024hecvl}. Early methods for phase recognition relied on hand-crafted visual features and temporal models like Hidden Markov Models (HMMs)~\cite{dergachyova2016automatic}. The emergence of deep learning techniques significantly improved phase recognition performance. Convolutional Neural Networks (CNNs) became the standard for extracting visual features from individual frames, while LSTMs and TCNs were employed to model temporal dependencies in video sequences~\cite{jin2020multi}.
Several studies have focused on effectively combining spatial and temporal information. SV-RCNet~\cite{jin2017sv} seamlessly integrates a ResNet and an LSTM network, enabling joint optimization of visual and temporal features during training.
Recognizing that surgical phases often exhibit long-range temporal dependencies, researchers have explored methods for incorporating longer temporal contexts. TMRNet~\cite{jin2021temporal} introduces a temporal memory relation network that utilizes a long-range memory bank and a temporal variation layer to encode long-range and multi-scale temporal information. TeCNO~\cite{czempiel2020tecno} employs a multi-stage TCN with dilated convolutions to achieve a larger receptive field for capturing longer temporal dependencies. Trans-SVNet~\cite{jin2022trans} leverages a hybrid embedding aggregation transformer to fuse spatial and temporal embeddings, allowing for the retrieval of critical information from longer temporal sequences.

Integrating visual data alone for surgical workflow recognition can be unreliable due to variations in surgical procedures, lighting, camera angles, and noise from blood or smoke~\cite{long2021relational}. Kinematics data, however, is more robust, capturing the position and orientation of instruments and providing valuable guidance for visual analysis. Recent methods, such as multimodal graph learning~\cite{liu2023visual,long2021relational} and feature fusion techniques~\cite{ahmidi2017dataset,qin2020temporal}, model and fuse relational features from both image and kinematics data. This approach allows models to learn useful representations of instrument tips unaffected by changes in the surgical scene. Developing robust models for surgical workflow recognition is challenging due to variations in surgical data from differing techniques, imaging equipment, and domain shifts. To address these challenges, multi-modal fusion can enhance model robustness, facilitating successful deployment in clinical settings. Developing robust models that can handle data variations and corrupted inputs is critical for successfully translating surgical workflow recognition technology into real-world operating rooms.

\subsection{Network Calibration}

Calibration in deep learning adjusts predicted probabilities to match true outcome likelihoods~\cite{wang2023calibration}. Deep neural networks often give overconfident predictions, affecting critical decisions. Methods to address this include post-hoc adjustments, regularization, and uncertainty estimation. 
Common post-hoc methods are Platt scaling~\cite{platt1999probabilistic}, Bayesian binning~\cite{naeini2015obtaining}, and isotonic regression~\cite{zadrozny2002transforming}. Recent work expands these, such as bias removal~\cite{huang2024post} and kernel density estimation~\cite{popordanoska2024beyond}.
Regularization techniques include L2~\cite{guo2017calibration}, entropy~\cite{pereyra2017regularizing}, and focal loss~\cite{lin2017focal}. New research explores implicit and explicit regularization~\cite{ijcai2024p499}, adaptive binning~\cite{berta2024classifier} and class-distribution-aware calibration~\cite{islam2021class}.
Uncertainty estimation uses Bayesian networks~\cite{blundell2015weight,fortunato2017bayesian}, ensembles~\cite{lakshminarayanan2017simple}, and Gumbel-softmax~\cite{jang2016categorical,pei2022transformer}. Recent advancements have further refined uncertainty estimation techniques. Kim et al.~\cite{kim2025uncertainty} replaces traditional softmax confidence scores with energy scores.
Proper calibration allows machine learning classifiers to deliver trustworthy and interpretable predictions, effectively aligning the gap between the model’s predictions and real probabilities.

Graph Neural Networks (GNNs) learn from graph-structured data, unlike traditional deep learning methods that handle independent data. Traditional calibration methods are ineffective for GNNs~\cite{teixeira2019graph}, which often show under-confidence~\cite{hsu2022makes, wang2022gcl}. Tailored calibration for GNNs is crucial for trustworthiness. Recent techniques include transforming logits to calibrated confidence for nodes~\cite{wang2021confident}, Graph Attention Temperature Scaling~\cite{hsu2022makes}, topology-aware calibration with cross-entropy and ECE~\cite{wang2024moderate}, Adaptive Spectral Temperature Scaling for local structures~\cite{xie2024exploring}, and analyzing decisive and homophilic edges~\cite{yang2024calibrating}.
Numerous studies highlight the susceptibility of deep learning to common data distortions such as blurring and Gaussian noise ~\cite{dodge2016understanding}. Recent research indicates that calibration methods, ensuring that predicted probabilities accurately reflect true outcomes, not only boost models’ accuracy but also enhance the robustness~\cite{park2024impact}. This alignment of predictions with reality is vital for maintaining performance and security in dynamic environments. Thus, the well-calibrated models provide trustworthy uncertainty estimates, making them less vulnerable to real-world noise attacks~\cite{ huang2024post, liu2024uncertainty}.

\section{Methodology}

\begin{figure*}[t]
    \centering
    \includegraphics[width=1.9\columnwidth]{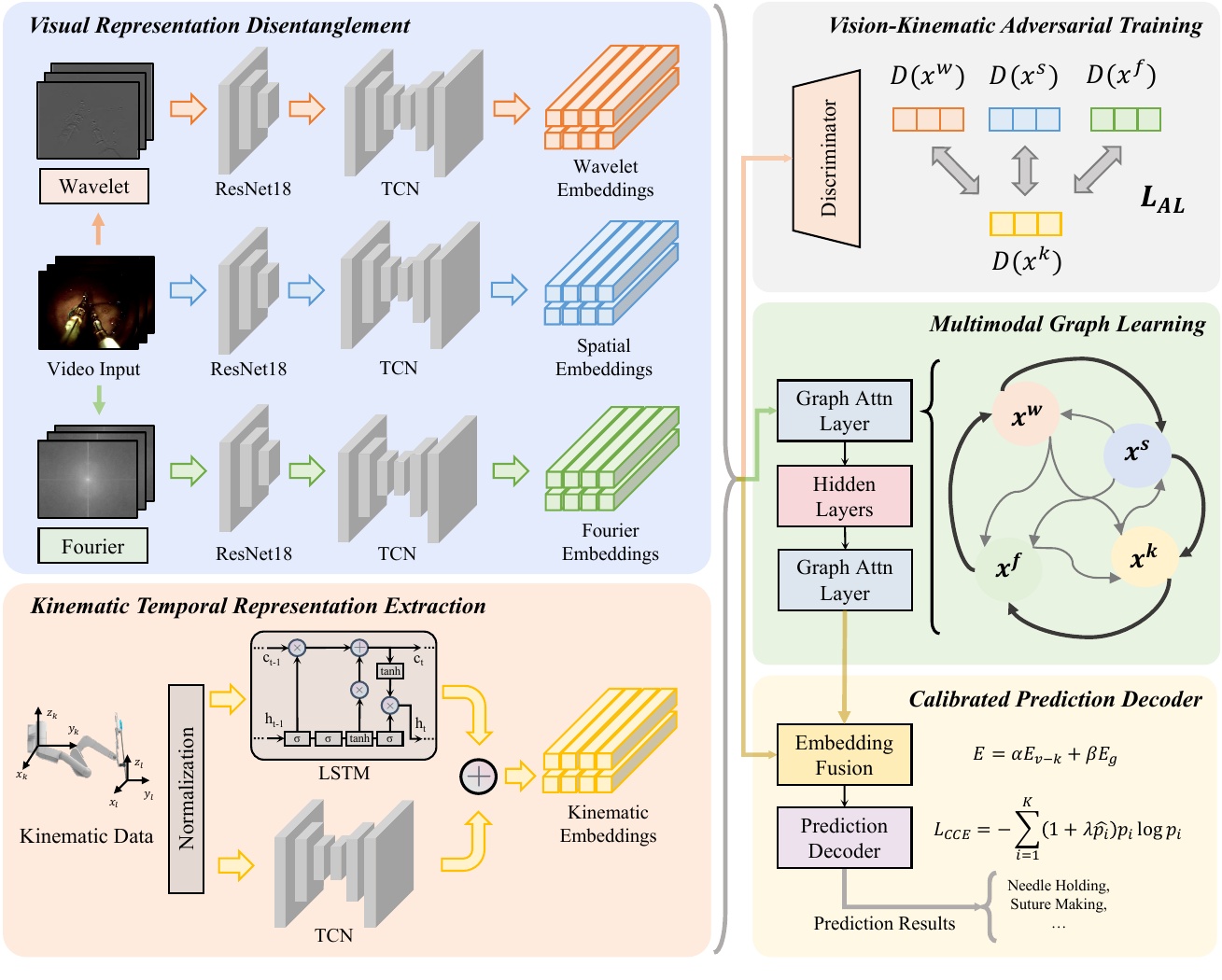}
    \caption{The overall architecture of our GRAD framework, consisting of the Multimodal Disentanglement Graph Network (Visual Representation Disentanglement module, the Kinematic Temporal Representation Extraction module, the Multimodal Graph Learning module), the Vision-Kinematic Adversarial Training, and the Calibrated Prediction Decoder. Our GRAD framework extracts visual and kinematic features through intra-modal feature mining. It uses adversarial learning to approximate the representation distributions of both. The graph network facilitates the interaction between the two modalities, and network calibration ensures robust prediction results.}
    \label{fig:framework}
\end{figure*}

\subsection{Preliminaries}
\subsubsection{Graph Network for Vision \& Kinematics}
Long et al.~\cite{long2021relational} dynamically integrate visual and kinematics information in the latent feature space of the relational graph convolutional network (RGCN). It first extracts embeddings from video sequences using temporal CNNs and from kinematics data of robotic arms using TCNs and LSTMs. These multimodal embeddings are then fed into a relational graph learning module, which identifies three relation types between nodes corresponding to video, left arm kinematics, and right arm kinematics. The module performs interactive message passing between nodes of different relation types through hierarchical graph convolutional layers. This allows the complementary information to be fused from visual and kinematics modalities through learning multiple relation-specific weights. Finally, the updated node representations are used for gesture classification through a prediction head.

\subsubsection{Graph Attention Network}
Using the masked self-attentional layers, graph attention networks (GATs)~\cite{velivckovic2017graph} allow assigning varying importance to different nodes without costly matrix operations or prior knowledge of the graph structure. Compared to other fusion methods, by learning the attention weights to be assigned, we can allocate appropriate focus on different modalities, thereby facilitating the integration and complementation of multimodal. The input is denoted as: ${\mathbf{h}}=\left\{ {\vec h_1,\vec h_2, \cdots,\vec h_N} \right\}$, $\vec h_i\in{\mathbb{R}^F}$, where $N$ denotes the number of nodes and $F$ represents the number of features associated with each node.

To obtain higher-level features, a shared linear transformation parametrized by a weight matrix, ${\bf{W}} \in {^{F' \times F}}$, is first applied to every node, after which self-attention is performed to compute attention coefficients, allowing each node to interact with every other node. The importance of node $j$’s feature to node $i$ shall be formulated as: 
\begin{equation}
    {e_{ij}} = attn({\mathbf{W}}{{\vec h}_i},{\mathbf{W}}{{\vec h}_j})
\end{equation}
where $attn:{\mathbb{R}^{F'}} \times {\mathbb{R}^{F'}} \to \mathbb{R}$ denotes a shared attentional mechanism. The ${e_{ij}}$ is only computed for nodes $j \in {{\bf{{\rm N}}}_i}$, where ${{\bf{N}}_i}$ is some neighborhood of node $i$ in the graph, thus, enhancing the model’s capacity to capture accurate relationships between various modalities.

\subsubsection{Adversarial Learning}

Adversarial learning usually comprises two main components: a generator and a discriminator. It was initially developed to tackle the generative modeling problem~\cite{goodfellow2020generative}. The generator aims to learn the underlying probability distributions and generate synthetic data based on the training samples. Conversely, the discriminator's objective is to differentiate between generated examples and real data. During training, the generator tries to minimize the difference between synthetic and real data, while the discriminator seeks to maximize the accuracy of distinguishing fake samples from real ones. This competitive dynamic helps the generator capture the true data distributions in the training examples. Given these characteristics, adversarial learning is well-suited for modality distribution translation, offering a promising approach for robust model training.

\subsection{Proposed Framework: GRAD}

Our GRAD framework, as presented in Figure~\ref{fig:framework}, comprises a visual representation learning branch with feature disentanglement, a kinematic feature extractor to extract sequential kinematics information, a graph attention network for multimodal fusion, the multimodal discriminator for adversarial learning, and the calibrated prediction decoder as the output module.

\subsubsection{Multimodal Disentanglement Graph Network}

We propose the Multimodal Disentanglement Graph Network (MDGNet) to encode and model the multimodal embeddings from vision and kinematics data.
First, we transform the kinematic signals into feature embeddings. The kinematic information is collected by the robotic platform, wherein each manipulator records seven degrees of freedom (DoFs): three for positional coordinates, three for orientation, and one for the gripper's opening angle. These values collectively describe the translation and rotation of a robotic manipulator in three-dimensional space. The task space can be presented as a 7-dimensional vector: ${\rm T} = [{x_l},{y_l},{z_l},{\alpha _l},{\beta _l},{\gamma _l},{\theta _l}]$, where $x_r$, $y_r$, $z_r$ are the Cartesian coordinates that specify the position of the left end effector, $\alpha _l$, $\beta _l$, $\gamma _l$ represent the left arm's orientation in term of roll, pitch, and yaw angles, $\theta_l$ denotes the opening angle of the left gripper. The same concept applies to the right robotic arm. Eventually, the kinematic information for two robotic arms at each frame is represented by 14 numerical values. The MISAW dataset further provides the output grip voltage for reference, so the kinematic information of the MISAW dataset is in the format of 16 numerical values. Extracting effective information from a single frame of kinematic data is challenging due to its sparse and complex nature, necessitating the modeling of temporal information over sequences. To preserve the fine-grained details and extract more temporally correlated information, we designed a temporal feature extractor that combines LSTM~\cite{graves2012long} with TCN~\cite{lea2017temporal} in parallel. By stacking a hierarchy of temporal convolutions, pooling, and upsampling operations, the TCN addresses the issue of independence of kinematic information over time, mitigating blurry and incoherent recognition. Using the multi-scale hierarchical structure, the TCN shall capture detailed relations between timestamps within the kinematics data. The LSTM maps the previous kinematics vector and previous hidden state to a new hidden state, capturing the long-range dependencies within the kinematic data. We then use the average of the LSTM and TCN outputs as the kinematic representations for subsequent multimodal fusion. The extraction process of the kinematic features $x_t^k,t \in \left\{ {1,2, \ldots ,T} \right\}$ is described as:
\begin{equation}
    x_t^k = \left[ {\operatorname{LSTM} (x_t) + \operatorname{TCN} (x_t)} \right] \times 0.5
\end{equation}
Consequently, we have achieved the modeling of kinematic information on a large temporal scale. We combine the left and right kinematic features for training, as our experiments demonstrate that this approach brings better performance compared to training them separately. 

Second, we disentangle the visual signal into spatial and frequency domains and extract feature representations. Understanding and interpreting visual data from surgical videos poses a significant challenge due to the rich and detailed information they contain. This includes various movements of instruments, subtle surgical procedures, and similar backgrounds throughout the videos. Distinguishing between different surgical phases within the same video can be highly complex. Current multimodal graph networks primarily consider the spatial domain while processing visual signals; however, the frequency domain can further preserve low-level statistical data~\cite{xu2021fourier} and modality-specific information~\cite{yang2020fda}. Therefore, to extract complex patterns in visual information that may not be discernible in raw visual data, we aim to provide explicit supervision of visual feature information by jointly mining the spatial and frequency domains. We further exploit the multi-level structural information and fundamental texture information of images through wavelet transform~\cite{bai2023llcaps} and Fourier amplitude spectra~\cite{zeng2024dbfft}, enabling the model to retain modality-invariant content and enhancing its robustness against corrupted inputs.

We employ wavelet and Fourier transforms for frequency domain analysis and feature extraction. Wavelet transforms capture image features at different scales, aiding the model in understanding the multi-level structure from coarse to fine details, and better recognizing local features such as edges, corners, and textures. They also exhibit high robustness to noise and variations under non-ideal conditions. The formula for extracting wavelet features is provided below:
\begin{equation}
\begin{split}
  & {A_j}(m,n) = \sum\limits_{w = 1}^{W} {\sum\limits_{h = 1}^{H} {x(h,w) \cdot {\phi _{j,m}}(h)} }  \cdot {\phi _{j,n}}(w) \hfill \\
  & D_j^{LZ}(m,n) = \sum\limits_{w = 1}^{W} {\sum\limits_{h = 1}^{H} {x(h,w) \cdot {\phi _{j,m}}(h)} }  \cdot \psi _{j,n}^Z(w) \hfill \\
  & D_j^{ZL}(m,n) = \sum\limits_{w = 1}^{W} {\sum\limits_{h = 1}^{H} {x(h,w) \cdot \psi _{j,m}^V(w) \cdot {\phi _{j,n}}(h)} }  \hfill \\
  & D_j^{ZZ}(m,n) = \sum\limits_{w = 1}^{W} {\sum\limits_{h = 1}^{H} {x(h,w) \cdot \psi _{j,m}^Z(w) \cdot \phi _{_{j,n}}^V(h)} }  \hfill \\ 
  & wavelet(m,n) \\
  & = [{A_j}(m,n)\parallel D_j^{LZ}(m,n)\parallel D_j^{ZL}(m,n)\parallel D_j^{ZZ}(m,n)]\\
\end{split}
\end{equation}
where the ${x(h,w)}$ represents the image in the spatial domain. ${A_j}(m,n)$, $D_j^{LZ}(m,n)$, $D_j^{ZL}(m,n)$ and $D_j^{ZZ}(m,n)$ denotes the approximation coefficients, horizontal detail coefficients, vertical detail coefficients and diagonal detail coefficients, respectively. Here, ${{\phi _{j,m}}}$, ${\psi _{j,m}^Z}$ and ${\psi _{j,m}^V}$ are the scaling function (father wavelet), horizontal and vertical wavelet functions (mother wavelet). 

Fourier amplitude spectra reveal the intensity of various frequency components in an image, assisting the model in understanding the fundamental textures and structural patterns while emphasizing global information for a better overall understanding of the image content. The formula for extracting Fourier amplitude spectra is provided below:
\begin{equation}
\begin{split}
    & F(p,q) = \sum\limits_{w = 1}^{W} {\sum\limits_{h = 1}^{H} {x(h,w){e^{ - 2\pi i\frac{{hp}}{H}}}{e^{ - 2\pi i\frac{{wq}}{W}}}} }\\
    & {F_S}(p,q) = {\left[ {{R^2}(p,q) + {I^2}(p,q)} \right]^{\frac{1}{2}}}
\end{split}
\end{equation}
where the ${x(h,w)}$ represents the spatial domain image and the $F{(p,q)}$ denotes each point in the Fourier domain. The discrete Fourier amplitude spectra are defined by the square root of the sum of the real and imaginary parts of $F{(p,q)}$. 

Specifically, for visual data in the spatial, wavelet, and Fourier domains, we establish the same feature extraction strategy. Given the spatial domain as an example: ${I_t},t \in \left\{ {1,2, \cdots ,T} \right\}$ denote the image frames from the video sequence. We first utilize the custom-trained ResNet-18~\cite{he2016deep} as the feature extraction backbone, which transforms $224 \times 224 \times 3$ RGB images into a spatial feature representation. Then, TCN is implemented to extract video features $x_t^i,t \in \left\{ {1,2, \cdots ,T} \right\}$ with long-range temporal patterns from a series of frame-wise features. The process can be formulated as:
\begin{equation}
    x_t^i = \operatorname{TCN} \left\{ {Dropout\left[ {\operatorname{ReLU} \left[ {\operatorname{CNN} ({I_t})} \right]} \right]} \right\}
\end{equation}
where a dropout rate of 0.5 is utilized to avoid overfitting.
The same visual feature extraction strategy is also applied in the wavelet and Fourier domains. The obtained feature representation can be represented as $x_t^w,t \in \left\{ {1,2, \cdots ,T} \right\}$ and $x_t^f,t \in \left\{ {1,2, \cdots ,T} \right\}$. The spatial, wavelet, and Fourier domains shall offer complementary and comprehensive image representations~\cite{strang1993wavelet,tan2024wavelet}. The spatial domain provides a direct representation of the original image. The wavelet domain enables multiresolution analysis, effectively capturing details and overall structures at different scales, making it particularly suitable for handling data with varying spatiotemporal characteristics. The Fourier domain, by analyzing the frequency components of an image, helps in understanding its global structure and patterns. By leveraging the complementary information from these domains and integrating global and local features, the model's ability to resist various types of noise and disturbances can be enhanced, further strengthening its robustness.

Finally, we establish our backbone with GAT~\cite{velickovic2017graph} to model the relational graph of visual and kinematic representations. Unlike previous methods that use stacked GCN layers~\cite{kipf2016semi,schlichtkrull2018modeling} to update features based on fixed adjacent matrix, which fail to grasp the inherent relationship between different modalities effectively, attention-based graph networks can learn the intrinsic interplay among diverse modalities and assign different importance levels to nodes corresponding to various modalities~\cite{vaswani2017attention}. This allows for better cross-modal information exchange as the model can discern certain inherent relationships dynamically during the learning process. 
The effectiveness of GAT is also verified in Section~\ref{exper:graph_network}. We consider the three visual domains (spatial, wavelet, Fourier) and the combined kinematic embedding as four sets of graph node features, denoted as ${\mathbf{n}} = \left\{ {\vec x_t^i,\vec x_t^w,\vec x_t^f,\vec x_t^k} \right\}$. These nodes are then refined through an aggregation process that incorporates messages from other nodes in the neighborhood. Taking spatial embeddings as an example, the formulation of the updated nodes $(\vec x_t^i)'$ is as follows:
\begin{equation}
\begin{split}
    (\vec x_t^i)' = \mathop {||}\limits_{k = 1}^K &  [ {\alpha _{ii}}{{\mathbf{W}}^i}\vec x_t^i + {\alpha _{iw}}{{\mathbf{W}}^w}\vec x_t^i \\
    & + {\alpha _{if}}{{\mathbf{W}}^f}\vec x_t^i + {\alpha _{ik}}{{\mathbf{W}}^k}\vec x_t^i ]
\end{split}
\end{equation}
where $K$ represents the number of independent attention mechanisms, and ${{\mathbf{W}}^ \cdot }$ denotes the weight matrix corresponding to the input linear transformation, ${\alpha _{iw}}$ denotes the attention score between nodes $i$ \& $w$, other attention scores can be inferred in the same way. To enable comparison of the attention coefficients across different nodes, the nodes in a neighborhood of $i$ are normalized using the softmax function. Taking ${\alpha _{iw}}$ as an example, the function is calculated:
\begin{equation}
\begin{split}
    & {\alpha _{iw}} = \operatorname{softmax} ({e_{iw}}) = \frac{{\exp \left( {{e_{iw}}} \right)}}{{\sum\nolimits_{j \in \left\{ {i,w,f,k} \right\}} {\exp \left( {{e_{ij}}} \right)} }}\\
    & {e_{iw}} = \operatorname{LeakyReLU} ({{{\mathbf{\vec a}}}^T}\left[ {{\mathbf{W}}{{\vec n}_i}||{\mathbf{W}}{{\vec n}_w}} \right])
\end{split}
\end{equation}
where ${{\mathbf{\vec a}}}$ is a weight vector and ${ \cdot ^T}$ represents its transposition. In this formula, ${\mathbf{W}} \in {\mathbb{R}^{F' \times F}}$ is a weight matrix where $F$ is the number of features in each node. By leveraging the attention mechanism in graph networks, the identification of critical associations can be enhanced while disregarding irrelevant information, thereby facilitating the capture of both intra-modal contextual information and inter-modal complementary information.

\subsubsection{Vision-Kinematic Adversarial Training}
Given that the visual and kinematic representations belong to different modalities, a significant disparity exists between their statistical properties. Previous study shows that matching those distributions in separate embedding spaces will significantly enhance the model’s sensitivity to irrelevant perturbations~\cite{mai2020modality,shang2023enhancing}. Therefore, mapping different representations into a common embedding space can reduce the gap between different modalities~\cite{liu2022target,wang2022adversarial}. In this case, we propose Vision-Kinematic Adversarial (VKA) learning to align them within a shared embedding space, which can facilitate the correlation of heterogeneous data from kinematic and visual modalities. Consequently, when facing random noise in real applications, the model can leverage information from alternative sources for compensation.

The multimodal adversarial learning implements a discriminator $D$ to distinguish the target modality as true but other source modalities as false, while the generator aims to fool the discriminator into classifying the target modality as true. The generator and discriminator engage in an adversarial, min-max game to learn a modality-invariant embedding space that allows diverse and complex modalities to be seamlessly matched. The loss function for this model can be broken down into two parts: fake adversarial loss ${{\mathop{\rm L}\nolimits} _{fal}}$ and the true adversarial loss ${{\mathop{\rm L}\nolimits} _{tru}}$. Through adding constraints to varying distributions, adversarial learning can learn common representations to effectively correlate heterogeneous information from diverse modalities. 

We have four inputs from two different modalities: $x_t^i $, $x_t^w$, and $x_t^f$ from the visual modality, and $x_t^k$ from the kinematic modality. We consider $x_t^k$ as the target modality, while the others are source modalities. The rationale of treating kinematic as the target modality has been empirically verified in Section~\ref{exper:abl_target}.
Visual information from the discrete wavelet transform and the Fourier transform provides important high-frequency and spectral details. Using these as additional source modalities can enhance adversarial learning to reduce the semantic gap between kinematic and visual representations. This helps constrain the differences between these modalities. While the generator turns distributions from different modalities into a shared embedding space, a discriminator distinguishes the representations originating from the kinematic or visual modalities. The discriminator is formulated as:
\begin{equation}
    D\left( {x_t^j} \right) = \operatorname{Sigmoid} \left\{ {l\left\{ {\operatorname{Tanh} \left[ {l\left[ {\operatorname{LeakyReLU} (l(x_t^j))} \right]} \right]} \right\}} \right\}
\end{equation}
where ${j \in \left\{ {i,w,f,k} \right\}}$ represents different modality, and $l\left(  \cdot  \right)$ denotes the linear function. In our case, the discriminator $D$ aims to differentiate the $x_t^k$ from kinematic modality as false but $x_t^i$, $x_t^w$, and $x_t^f$ from visual modality as true. The loss function of adversarial learning is defined as:
\begin{equation}
\begin{split}
    & \mathcal{L} _{AL} = \bigg[{\mathcal{L} _{fal}}(x_t^k) + {\mathcal{L} _{tru}}(x_t^i,x_t^w,x_t^f)\bigg]\times0.5, \\
    & \mathcal{L} _{fal} =  \log \left( {1 - D \left( {x_t^k} \right)} \right), \\
    & \mathcal{L} _{tru} =  \log \left( {D \left( {x_t^i} \right)} \right) + \log \left( {D \left( {x_t^w} \right)} \right) + \log \left( D \left( {x_t^f} \right) \right)
\end{split}
\end{equation}
When the discriminator cannot accurately distinguish kinematic representations from visual ones, it indicates that the learned distributions are mapped into a modality-invariant embedding space. By aligning the distributions of the kinematic modality with those of the visual modality, we can maintain enhanced and robust modality representations. Through adversarial learning, our GRAD framework better implements cross-modal information complementarity. Thus, when visual signals encounter corruption, the model's sensitivity to random noise is reduced, improving the robustness of the surgical workflow recognition process.

\subsubsection{Contextual Calibrated Decoder}

We aim to enhance the robustness of our multimodal graph network against corrupted data by proposing our contextual calibrated decoder, which consists of heterogeneous embedding fusion and calibrated optimization fusion. Typically, the depth of deep neural networks helps improve representation capabilities but can lead to overfitting~\cite{guo2017calibration}. However, GNNs usually have a relatively low capacity because stacking more GNN layers can cause gradient vanishing issues. Previous work~\cite{wang2022gcl} has verified that although the accuracy of GNNs decreases with more layers, the confidence score first rises and then falls. This suggests that the low confidence of GNNs might be due to their shallow layers. In this case, we explore the heterogeneous embedding fusion setup to integrate the vision-kinematic feature embeddings and the graph output embeddings, and feed the integrated embeddings to the prediction encoder. The vision-kinematic embeddings can be expressed as:
\begin{equation}
    {E_{v - k}} = l\left( {x_t^i\parallel x_t^w\parallel x_t^f\parallel x_t^k} \right)
\end{equation}

The graph output embeddings can be expressed as:
\begin{equation}
    {E_g} = l\left[ {GAT\left( {x_t^i,x_t^w,x_t^f,x_t^k} \right)} \right]
\end{equation}
where $l\left(  \cdot  \right)$ represents the linear function. Then, the prediction encoder input is formulated as:
\begin{equation}
    E = \alpha E_{v-k} + \beta E_{g}
    \label{equ:feature_output}
\end{equation}
in which the $\alpha$ and $\beta$ are the combination ratios of the vision-kinematic embeddings and the graph embeddings. The two coefficients are empirically verified through the experiments in Section~\ref{exper:abl_embedding_fusion}. 
By combining and interacting vision-kinematic features with GNNs embeddings, we can achieve better feature interaction and complementarity, leading to improved prediction results. In the following experiment section, we perform grid-based ablation studies on the parameters $\alpha$ and $\beta$.
Our prediction encoder comprises a fully connected layer, and the classification nodes correspond to the number of classes.

As for the optimization function, the typical cross-entropy loss will be applied to the decoder to optimize the network. However, we aim to achieve calibration by amplifying the output logits during the training of graph networks. For a model $\mathcal{G}$, we set $\{f, y\}$ are the pair of the input feature and the ground truth label, and $\hat{p}_{y}=\mathcal{G}\left(f, y\right)$ is the output probability that $\mathcal{G}$ predicts a label $y$ for an input feature $f$. The predicted confidence is $\hat{p}=\max \hat{p}_{y}$, and the label that $\mathcal{G}$ predicts is $\hat{y}=\operatorname{argmax} \hat{p}_{y}$. 
The GNN is considered perfectly calibrated when the predicted probability, $\hat{p}$, aligns exactly with $p$, the true probability of making a correct prediction for node $i$.

The conventional cross-entropy loss can be expressed as:
\begin{equation}
    \mathcal{L}_{CE} = -\sum_{i=1}^K p_i \log \hat{p}_i
\end{equation}
where $p$ is the target distribution and $\hat{p}$ is predicted distribution. The relationship between the cross-entropy loss and the KL divergence can be represented as:
\begin{equation}
    D_{KL}(P \parallel \hat{P}) = \mathcal{H}(P) - \mathcal{L}_{CE}
\end{equation}
In which $\mathcal{H}(P) = -\sum_{i=1}^K p_i \log p_i$ is a constant, representing the entropy for a given real distribution. We aim to amplify the logits to increase the model's confidence during the computation of max. Therefore, we add a minimal-entropy regularization term to the implicit minimization performed by cross-entropy loss. We intend to incorporate a regularization term into the cross-entropy loss such that it minimizes when the confidence equals the true probability. Assume the regularization term is $P \mathcal{H}(\hat{P})$, meaning the target probability distribution multiplied by the entropy of the predicted distribution. If the predicted distribution aligns with the target distribution, the term will lead to lower entropy and a smaller value for the given expression. By converging this regularization term, the model can achieve better calibration, resulting in stronger robustness. With the new regularization term, the new loss function can be formulated as:
\begin{equation}
\begin{split}
    \mathcal{L}_{CCE} & = \mathcal{L}_{CE} - \lambda P \mathcal{H}(\hat{P}) \\
    & = -\sum_{i=1}^K p_i \log \hat{p}_i - \lambda \sum_{i=1}^K {p_i \hat{p}}_i \log \hat{p}_i \\
    & = -\sum_{i=1}^K\left(1+ \lambda \hat{p}_i\right) p_i \log \hat{p}_i
    \label{equ:ca_loss}
\end{split}
\end{equation}
$\lambda$ is set to 0.02, which is proven to be the most effective parameter by 
our experiments. Therefore, our final loss function can be expressed as:
\begin{equation}
    \mathcal{L} = \gamma \mathcal{L}_{CCE} + \delta \mathcal{L}_{AL}
    \label{equ:loss}
\end{equation}
$\gamma$ and $\delta$ are the loss ratios, which are also empirically confirmed through the experiments in Section~\ref{exper:abl_loss}.

\section{Experiments}
\subsection{Datasets}
Extensive experiments are carried out using two publicly available datasets for surgical workflow recognition: the MIcro Surgical Anastomose Workflow (MISAW) dataset and the CUHK-MRG dataset, which is a multimodal dataset created on the da Vinci Research Kit platform at The Chinese University of Hong Kong (CUHK), as described in the paper MRG-Net~\cite{long2021relational}. These two datasets originate from two different types of robot-assisted surgeries, and include visual data, kinematic data, and workflow recognition annotations. Prior works have validated their proposed methods on these datasets~\cite{long2021relational,yamada2024multimodal}.

\begin{figure*}[t]
    \centering
    \includegraphics[width=1\linewidth]{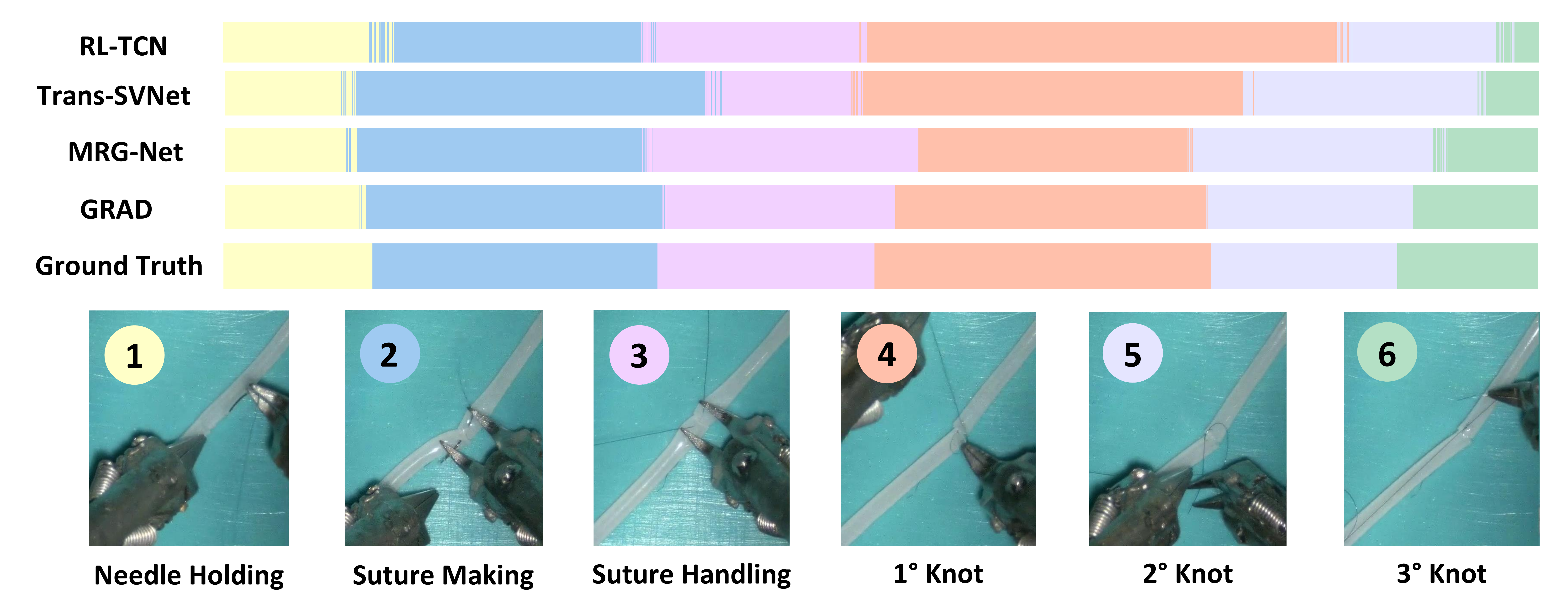}
        \caption{Visualization of GRAD compared to MRG-Net (Vis+Kin), Trans-SVNet (Vis), and RL-TCN (Kin) model on MISAW dataset. The different color bands represent each category, which are needle holding, suture making, suture handling, $1^ \circ$ knot, $2^ \circ$ knot, and $3^ \circ$ knot (with order).}
    \label{fig:misaw}
\end{figure*}

\begin{figure*}[t]
    \centering
    \includegraphics[width=1\linewidth]{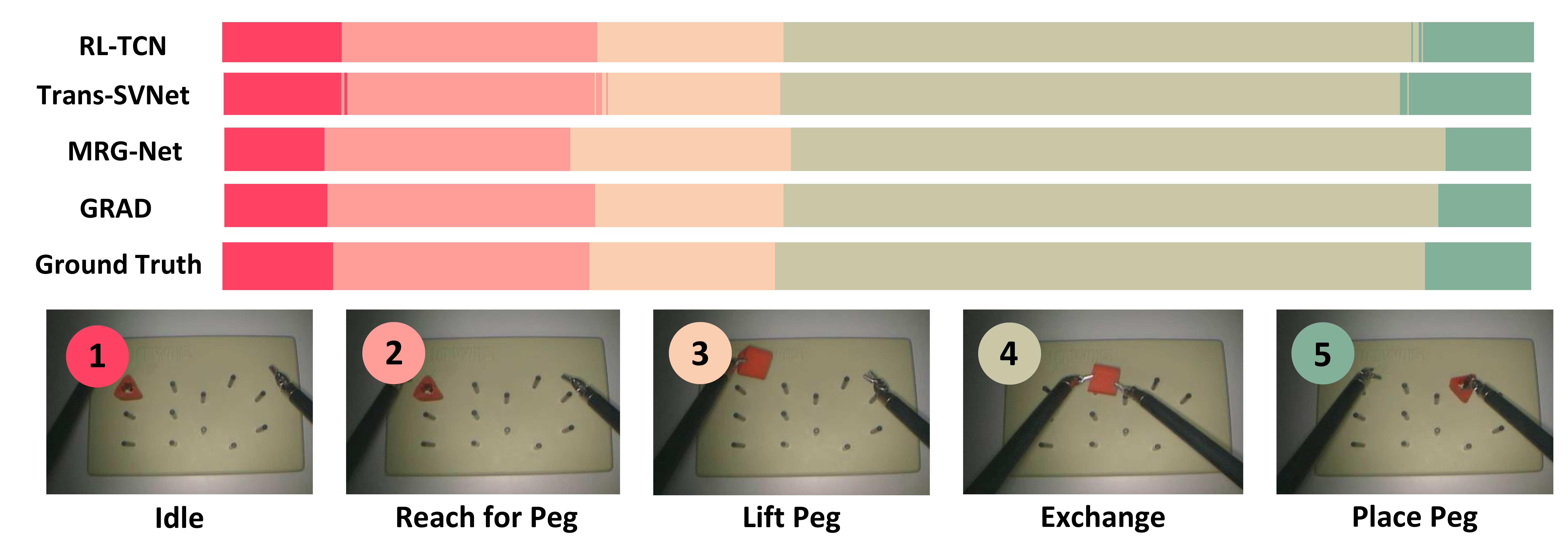}
    \caption{Figure of CUHK-MRG Dataset illustrates five different gesture steps, which are idle (no action performed), reach for peg (with left hand), lift peg (with left hand), exchange (transfer the peg to right hand), place peg (with right hand).}
    \label{fig:CUHK-MRG}
\end{figure*}

\textbf{MISAW Dataset}~\cite{huaulme2021micro}: The MISAW dataset consists of 27 recordings of micro-surgical anastomosis procedures conducted on synthetic blood vessels, featuring the work of three surgeons and three engineering students. The kinematic data and video footage were recorded concurrently at 30 Hz via a high-resolution stereo microscope, which has a resolution of 960 by 540 pixels, along with a master-slave robotic system. Each individual frame is tagged with six categories of surgical gestures, as illustrated in Figure~\ref{fig:misaw}. The dataset was divided into a training set with 17 cases and a testing set with 10 cases.

\textbf{CUHK-MRG Dataset}~\cite{long2021relational}: This dataset consists of 24 sequences and the five different gesture steps, which can be seen from Figure~\ref{fig:CUHK-MRG}. The kinematic data encompasses the position and orientation of the end-effector and the gripper's opening angle. At the same time, video recordings are made at a frequency of 30 Hz. The experimental setting adopts 3-fold cross-validation (8 sequences for training and 4 for testing) following~\cite{long2021relational}.

\begin{table*}
    \centering
    \caption{Comparison of our proposed method against the state-of-the-art single-modality and multi-modality solutions on the MISAW dataset. $*$ indicates that the model shall shuffle the data during training, making it infeasible to calculate the edit score. $\dagger$ means results are extracted from~\cite{huaulme2021micro}.}
    \setlength{\tabcolsep}{10pt} 
    \resizebox{\textwidth}{!}{
    \begin{tabular}{l|cc|cccccccc}
         \hline
         \multicolumn{1}{c|}{Methods} & Vis & Kin &  Acc &  Edit &  OP &  OR &  OF1 &  CP&  CR& CF1 \\ \hline
         LSTM~\cite{hochreiter1997long} & \scriptsize{\XSolidBrush} & \scriptsize{\CheckmarkBold} & 42.65&  42.14&  42.39&  44.57&   43.59&  36.78&  42.78&  38.87\\
         GRU~\cite{cho2014properties} & \scriptsize{\XSolidBrush} & \scriptsize{\CheckmarkBold} &  61.63&  59.62&  72.01&  68.87&  70.40&  38.47&  41.79& 40.06\\
         TCN~\cite{lea2016temporal} & \scriptsize{\XSolidBrush} & \scriptsize{\CheckmarkBold}&  60.54&  44.53&  60.54&  60.52&  60.52&  61.23&  57.13&  52.51\\
         RL-TCN~\cite{liu2018deep} & \scriptsize{\XSolidBrush} & \scriptsize{\CheckmarkBold} &  67.72&  64.36&  68.97&  72.42&  70.65&  61.97&  69.73& 65.63\\ \hline
         GRU~\cite{cho2014properties} & \scriptsize{\CheckmarkBold} & \scriptsize{\XSolidBrush} &  71.12&  70.32&  68.05&  73.87&  70.84&  46.20&  48.50& 47.32\\ 
         TCN~\cite{lea2016temporal} & \scriptsize{\CheckmarkBold} & \scriptsize{\XSolidBrush} &  68.16&  77.07&  68.16&  69.02&  68.59&  56.64&  60.61& 59.38\\
         RL-TCN~\cite{liu2018deep} & \scriptsize{\CheckmarkBold} & \scriptsize{\XSolidBrush} &  68.63&  79.02&  66.30&  67.70&  66.99&  59.15&  63.19& 61.10\\
         TeCNO~\cite{czempiel2020tecno} & \scriptsize{\CheckmarkBold} & \scriptsize{\XSolidBrush} &  76.64&  77.26&  73.73&  77.68&  75.65&  61.50&  64.00& 62.73\\ 
         TMRNet~\cite{jin2021temporal} & \scriptsize{\CheckmarkBold} & \scriptsize{\XSolidBrush} &  74.44&  79.10&  74.02&  74.71&  74.36&  63.12&  60.45& 61.76\\
         Trans-SVNet~\cite{jin2022trans} & \scriptsize{\CheckmarkBold} & \scriptsize{\XSolidBrush} &  77.34&  78.60&  74.28&  76.91&  75.58& 61.55&  62.15&  61.85\\ 
         AI-Endo~\cite{cao2023intelligent} & \scriptsize{\CheckmarkBold} & \scriptsize{\XSolidBrush} &  70.35&  76.09&  74.28&  75.34&  74.81&  67.73&  70.87& 69.27\\
         Surgformer~\cite{yang2024surgformer} & \scriptsize{\CheckmarkBold} & \scriptsize{\XSolidBrush} &  56.25&  $*$&  40.96&  40.48&  40.73&  36.51&  47.01& 41.10\\
         Pitfall-BN~\cite{rivoir2024pitfalls} & \scriptsize{\CheckmarkBold} & \scriptsize{\XSolidBrush} &  71.07&  $*$&  71.13&  71.06&  71.09&  71.49&  63.89& 67.48\\ \hline
         LC-SC-CRF~\cite{lea2016learning} & \scriptsize{\CheckmarkBold} & \scriptsize{\CheckmarkBold} &  78.62&  87.74&  78.76&  78.25&  78.51&  77.48&  78.86& 78.14 \\
         MRG-Net$\dagger$~\cite{long2021relational} & \scriptsize{\CheckmarkBold} & \scriptsize{\CheckmarkBold} &  83.75&  -&  -&  -&  -& \textbf{86.74}&  81.34&  80.97 \\
         GRAD (\textbf{Ours}) & \scriptsize{\CheckmarkBold} & \scriptsize{\CheckmarkBold} & \textbf{86.87} & \textbf{97.50}& \textbf{86.83}&  \textbf{88.23}& \textbf{87.52}& 83.07& \textbf{91.66}& \textbf{87.16} \\\hline
    \end{tabular}}
    \label{table:misaw}
\end{table*}

\subsection{Implementation Details}
The end-to-end framework comprises separate visual and kinematic feature extractors, multimodal graph representation learning, adversarial discriminators, and the workflow prediction decoder. Initially, the video frames are resized to a resolution of $320 \times 256$ pixels, with a random crop to 224 by 224 pixels applied to minimize training load and mitigate the risk of overfitting. The video sequence information is then trained through a pre-trained ResNet18~\cite{he2016deep}. After that, the spatial-CNN features are generated from the backbone~\cite{lea2017temporal} to make the training procedure more efficient. 
The graph attention layers feature hidden and output states that are 64 dimensions in size, with a dropout rate of 0.5 implemented. The discriminators are composed of three fully connected layers with dimensions $\left\{ {64, 64} \right\}$, $\left\{ {64, 16} \right\}$, $\left\{ {16, 1} \right\}$. The workflow prediction decoder is also a fully connected layer, and its output dimension corresponds to the number of categories in the dataset. The network is optimized with the Adam optimizer, a learning rate of $1{\times 10^{-4}}$, and a batch size of $64$.

\begin{table*}
    \centering
    \caption{Comparison of our proposed method against the state-of-the-art single-modality and multi-modality solutions on the CUHK-MRG dataset. $*$ indicates that the model shall shuffle the data during the training process, making it unfeasible to calculate the edit score.}
    \setlength{\tabcolsep}{10pt} 
    \resizebox{\textwidth}{!}{
    \begin{tabular}{l|cc|cccccccc}
         \hline
         \multicolumn{1}{c|}{Methods} & Vis & Kin & Acc &  Edit &  OP &  OR &  OF1 &  CP&  CR& CF1 \\ \hline
         LSTM~\cite{hochreiter1997long}& \scriptsize{\XSolidBrush} & \scriptsize{\CheckmarkBold} &  81.59 & \textbf{100.00}&  86.41&  85.44&  85.90&  86.85&  88.28&  87.54\\
         GRU~\cite{cho2014properties} & \scriptsize{\XSolidBrush} & \scriptsize{\CheckmarkBold} &  86.27&  \textbf{100.00}&  90.11&  89.45&  89.77&  86.72&  88.21& 87.45\\
         TCN~\cite{lea2016temporal} & \scriptsize{\XSolidBrush} & \scriptsize{\CheckmarkBold} & 91.29&  \textbf{100.00}&  92.46&  91.41&  91.93&  89.10&  89.23&  88.82\\
         RL-TCN~\cite{liu2018deep} & \scriptsize{\XSolidBrush} & \scriptsize{\CheckmarkBold} &  89.35&  97.62&  91.44&  91.93&  91.67&  88.03&  89.29& 88.64\\ \hline
         GRU~\cite{cho2014properties} & \scriptsize{\CheckmarkBold} & \scriptsize{\XSolidBrush} &  86.80&  \textbf{100.00}&  89.79&  88.79&  89.27&  86.49&  87.73& 87.08\\
         TCN~\cite{lea2016temporal} & \scriptsize{\CheckmarkBold} & \scriptsize{\XSolidBrush} &  90.40&  97.22&  92.58&  91.85&  92.21&  88.85&  88.68& 88.73\\
         RL-TCN~\cite{liu2018deep} & \scriptsize{\CheckmarkBold} & \scriptsize{\XSolidBrush} &  89.56&  \textbf{100.00}&  91.96&  92.24&  92.10&  88.31&  89.12& 88.70\\
         TeCNO~\cite{czempiel2020tecno} & \scriptsize{\CheckmarkBold} & \scriptsize{\XSolidBrush} &  91.00&  91.31&  90.55&  91.37&  90.96&  86.13&  85.82& 85.96\\
         TMRNet~\cite{jin2021temporal} & \scriptsize{\CheckmarkBold} & \scriptsize{\XSolidBrush} &  90.74&  89.29&  87.50&  87.56&  87.53&  84.83&  75.72& 79.97\\ 
         Trans-SVNet~\cite{jin2022trans} & \scriptsize{\CheckmarkBold} & \scriptsize{\XSolidBrush} &  89.69&  90.02&  89.47&  89.24&  89.35&  84.72&  84.68& 84.70\\
         AI-Endo~\cite{cao2023intelligent} & \scriptsize{\CheckmarkBold} & \scriptsize{\XSolidBrush} &  89.96&  88.18&  89.19&  86.09&  87.61&  75.38&  74.21& 74.79\\
         Surgformer~\cite{yang2024surgformer} & \scriptsize{\CheckmarkBold} & \scriptsize{\XSolidBrush} &  90.25&  $*$&  92.18&  89.80&  90.51&  84.83&  87.72&86.25\\
         Pitfall-BN~\cite{rivoir2024pitfalls} & \scriptsize{\CheckmarkBold} & \scriptsize{\XSolidBrush} &  78.25&  $*$& 77.71&  76.88&  77.29& 78.66&  77.68 & 78.11\\ \hline
         LC-SC-CRF~\cite{lea2016learning} & \scriptsize{\CheckmarkBold} & \scriptsize{\CheckmarkBold} &  89.97&  92.91&  89.85&  90.03&  89.94&  88.89&  89.50& 89.12\\
         MRG-Net~\cite{long2021relational} & \scriptsize{\CheckmarkBold} & \scriptsize{\CheckmarkBold} &  89.86&  \textbf{100.00}&  92.80&  92.24&  92.53&  88.88&  89.41& 89.12\\
         GRAD (\textbf{Ours})  & \scriptsize{\CheckmarkBold} & \scriptsize{\CheckmarkBold}&  \textbf{92.38}& \textbf{100.00}& \textbf{92.83}&  \textbf{93.20}&  \textbf{93.01}& \textbf{91.40}&  \textbf{92.60}& \textbf{92.00}\\\hline
    \end{tabular}}
    \label{table:cuhk}
\end{table*}

\begin{table*}
    \centering
    \caption{Ablation study of the proposed modules on the MISAW Dataset. `CAL' denotes the calibration module, `VRD' denotes visual representation disentanglement, and `VKA' denotes vision-kinematic adversarial training. We (i) exclude the calibration module, (ii) remove the visual feature disentanglement, and (iii) remove the vision-kinematic adversarial training.}
    \setlength{\tabcolsep}{12pt} 
    \resizebox{\textwidth}{!}{
    \begin{tabular}{c|c|c|cccccccc}
         \hline
         CAL &  VRD &  VKA &  Acc &  Edit &  OP &  OR &  OF1 &  CP &  CR & CF1 \\ \hline
        \XSolidBrush&  \XSolidBrush&  \XSolidBrush&  83.75&  90.29&  84.75&  84.56&  84.66&  81.00&  84.75& 82.83\\
         \Checkmark&  \XSolidBrush&  \XSolidBrush&  82.12&  94.58&  84.85&  86.64&  85.74&  83.83&  89.39& 86.52\\
         \XSolidBrush&  \Checkmark&  \XSolidBrush&  84.02&  97.29&  85.72&  87.72&  86.71&  84.53&  87.60& 86.04\\
         \XSolidBrush&   \XSolidBrush&  \Checkmark& 83.01&  91.39&  83.93&  86.77&  85.33&  82.59&  85.75& 84.14\\
         \Checkmark&  \Checkmark&  \XSolidBrush&  82.71&  96.25&  85.83&  87.90&  86.85&  \textbf{84.66}&  89.92& \textbf{87.21}\\
         \XSolidBrush&  \Checkmark&  \Checkmark&  84.42&  \textbf{97.50}&  86.20&  87.89&  87.04&  82.67&  86.83& 84.67\\
         \Checkmark&  \XSolidBrush&  \Checkmark&  84.20&  \textbf{97.50}&  85.56&  87.90&  86.71&  81.77&  90.80& 86.05\\
         \Checkmark&  \Checkmark&  \Checkmark&  \textbf{86.87}&  \textbf{97.50}&  \textbf{86.83}&  \textbf{88.23}&  \textbf{87.52}&  83.07&  \textbf{91.66}& 87.16\\\hline
         
    \end{tabular}}
    \label{table:component}
\end{table*}

The employed evaluation metrics on those two datasets include (i) Accuracy (\%) in the frame-wise level, referring to the calculation of the proportion of frames that are correctly identified; (ii) Edit Score~\cite{lea2016segmental} (ranging from 0 to 100, with higher values indicating better performance), which is intended to evaluate effectiveness in video segmentation, specifically focusing on maintaining temporal smoothness. For the ground truth segments $S = S_1, S_2, \dots, S_x$ and the predicted segments $S' = S'_1, S'_2, \dots, S'_y$, the edit number $N_{edit}$ is normalized by the maximum value of $x$ and $y$. Then, the $\mathrm{Edit\ Score}$ is computed as $\mathrm{Edit\ Score} = 100 \times (1 - N_{edit})$; (iii) Overall Precision (OP),  is calculated as the proportion of true positives (correctly identified positive instances) to the total of true positives and false positives (incorrectly identified positive instances); (iv) Overall Recall (OR), represents the proportion of true positives compared to the total number of true positives and false negatives (the positive instances that were misclassified as negatives); (v) Overall F1 Scores (OF1), the harmonic mean of precision and recall; (vi) average per-Class Precision (CP), is the average of precision values for each class, vii) average per-Class Recall (CR), is the average of recall values for each class; (viii) average per-Class F1 Scores (CF1), is the average of F1 scores for each class. Our method is compared against the state-of-the-art kinematics-based, vision-based, and multimodal solutions. 

\begin{table*}
    \centering
    \caption{Ablation study of the joint modeling of the left and right kinematic data on the MISAW dataset. The first row indicates that the method shall learn the kinematic features of the left and right hands separately and model them as individual graph nodes. The second row indicates that the method shall learn the kinematic features of both hands jointly and model them as a combined graph node.}
    \setlength{\tabcolsep}{14pt} 
    \resizebox{\textwidth}{!}{
    \begin{tabular}{c|cccccccc}
        \hline
        \makecell{Joint Modeling of \\ Left-Right Kinematic} &  Acc &  Edit &  OP &  OR &  OF1 &  CP &  CR & CF1 \\ \hline
        \XSolidBrush & 83.43 & 97.50 & 86.50 & 85.79 & 86.14 & 75.30 & 80.55 & 77.85 \\
        \Checkmark & \textbf{86.87}&  \textbf{97.50}&  \textbf{86.83}&  \textbf{88.23}&  \textbf{87.52}&  \textbf{83.07}&  \textbf{91.66} & \textbf{87.16}\\\hline
         
    \end{tabular}}
    \label{table:left_right}
\end{table*}

\begin{table*}
    \centering
    \caption{Ablation study for the effectiveness of the wavelet and Fourier domain in the graph disentanglement on the MISAW dataset.}
    \setlength{\tabcolsep}{12pt} 
    \resizebox{\textwidth}{!}{
    \begin{tabular}{c|c|cccccccc}
         \hline Wavelet &  Fourier&  Acc &  Edit &  OP &  OR &  OF1 &  CP &  CR & CF1 \\ \hline
         \XSolidBrush&  \XSolidBrush& 84.20& \textbf{97.50}&  85.56&  87.90&  86.71&  81.77&  90.80& 86.05\\
         \XSolidBrush&  \Checkmark&  84.12&  \textbf{97.50}&  83.84&  85.32&  84.57&  80.01&  87.73& 83.69\\
         \Checkmark&  \XSolidBrush&  84.36&  95.00&  86.54&  87.10&  86.82&  82.96&  87.85& 85.33\\
         \Checkmark&  \Checkmark&   \textbf{86.87}&  \textbf{97.50}&  \textbf{86.83}&  \textbf{88.23}&  \textbf{87.52}&  \textbf{83.07}&  \textbf{91.66}& \textbf{87.16}\\\hline
    \end{tabular}}
    \label{table:des}
\end{table*}

\subsection{Experimental Results}
\subsubsection{Comparison with the State-of-the-Art}

We compare our proposed GRAD framework against various state-of-the-art surgical workflow recognition methods, including (i) kinematic-based methods: LSTM~\cite{hochreiter1997long}, GRU~\cite{cho2014properties},  TCN~\cite{lea2016temporal}, RL-TCN~\cite{liu2018deep}; (ii) vision-based methods: GRU~\cite{cho2014properties}, TCN~\cite{lea2016temporal}, RL-TCN~\cite{liu2018deep}, TeCNO~\cite{czempiel2020tecno}, TMRNet~\cite{jin2021temporal}, Trans-SVNet~\cite{jin2022trans}, AI-Endo~\cite{cao2023intelligent}, Surgformer~\cite{yang2024surgformer}, Pitfall-BN~\cite{rivoir2024pitfalls}; (iii) multimodal methods: LC-SC-CRF~\cite{lea2016learning}, MRG-Net~\cite{long2021relational}.

\begin{table*}
    \centering
    \caption{Ablation study for the effectiveness of different graph networks on the MISAW dataset. `GCN' represents the graph convolutional networks. `RGCN' denotes the relational graph convolutional network. `GAT' is the graph attention network.}
    \setlength{\tabcolsep}{12pt} 
    \resizebox{\textwidth}{!}{
    \begin{tabular}{c|cccccccc}
         \hline Graph Networks &  Acc &  Edit &  OP &  OR &  OF1 &  CP &  CR & CF1 \\ \hline
         GCN~\cite{kipf2016semi}& 84.79& 94.29&  84.56&  86.06&  85.30&  \textbf{85.02}& 89.86& 85.67\\
         RGCN~\cite{schlichtkrull2018modeling}&  83.42&  94.58&  \textbf{86.91}&  87.72&  87.31&  84.47&  88.18& 84.63\\
         GAT~\cite{velivckovic2017graph}&   \textbf{86.87}&  \textbf{97.50}&  86.83&  \textbf{88.23}&  \textbf{87.52}& 83.07&  \textbf{91.66}& \textbf{87.16}\\\hline
    \end{tabular}}
    \label{table:gcn}
\end{table*}

For the MSIAW dataset, based on the experimental results in Table~\ref{table:misaw}, the model GRU using visual data presents better results on most indicators compared to the same model using motion data where the accuracy is increased by 9.49\%. The same phenomenon can be observed in TCN and RL-TCN models. Moreover, the highest score on accuracy with single-modality is 77.34\% achieved by the Trans-SVNet using visual data. This accuracy surpasses the highest attainable accuracy using motion data by 9.62\%. It demonstrates that the utilization of visual data achieves superior results compared to the model trained on motion data. This signifies that video encompasses more information than the kinematic data on this dataset and can facilitate the model’s training process and enhance accuracy. By comparing the performance of models using multimodal data and unimodal data, it can be observed that using multimodal data can increase the performance on most metrics by at least 2\% compared to unimodal data. Moreover, as can be seen, the LC-SC-CRF model can improve the edit score by 8.72\% which signifies that with multimodal data, the model can be trained more thoroughly, leading to a much better result. Therefore, it can be inferred that multimodal data has the ability to integrate information from two unimodal data sources and offer more comprehensive knowledge for phrase recognition. 

Furthermore, on the MISAW dataset, our model achieves 86.87\% of accuracy and outperforms the second-highest score by 3.12\%. Our model also presents 97.50\% on the edit score, suggesting our superiority in learning the sequential order. Generally, our model yields 86.83\% on the overall precision score, 88.23\% on the overall recall score, and 87.52\% on the overall F1 score, which increases the second-highest score by at least 8\% and presents the highest score on average per-class recall and F1 scores. 
However, the performance at the average per-class precision metric is lower than MRG-Net, showcasing the limitation of our proposed method. Even though our model learns the ordering of actions well, it gets confused in discerning the beginning and end of certain phrases. Thus, it leads to lower precision scores on certain categories and decreases the average per-class scores. Nevertheless, our method achieves the best performance on most metrics. Overall, from the analysis of the comparison results, it can be observed that our model showcases superior performance against the existing state-of-the-art methods. Besides, Figure~\ref{fig:misaw} presents the visualization and qualitative comparison of the MISAW dataset.

\begin{table*}
    \centering
    \caption{Ablation study of the combination ratio of the vision-kinematic embeddings and graph embeddings on the MISAW dataset, e.g., the $\alpha$ and $\beta$ in the Equation~(\ref{equ:feature_output}).}
    \setlength{\tabcolsep}{12pt} 
    \resizebox{\textwidth}{!}{
    \begin{tabular}{c|c|cccccccc}
         \hline
         V-K ($\alpha$) &  Graph ($\beta$) &  Acc &  Edit &  OP &  OR &  OF1 &  CP &  CR & CF1 \\ \hline
         0.0&  1.0&  85.16&  97.14&  83.81&  84.35&  84.08&  79.40&  85.52& 82.35\\
         0.1&  0.9&  85.53&  \textbf{97.50}&  82.08&  84.19&  83.12&  78.51&  87.80& 82.90\\
         0.2&  0.8&  85.38&  \textbf{97.50}&  83.23&  84.84&  84.03&  79.71&  89.22& 84.20\\
         0.3&  0.7&  86.87&  \textbf{97.50}&  \textbf{86.83}&  88.23&  \textbf{87.52}&  \textbf{83.07}&  \textbf{91.66}& \textbf{87.16}\\
         0.4&  0.6&  84.38&  95.00&  81.38&  83.87&  82.60&  76.77&  84.86& 80.61\\
         0.5&  0.5&  83.87&  95.00&  83.46&  84.68&  84.07&  78.44&  85.73& 81.93\\
         0.6&  0.4&  83.89&  \textbf{97.50}&  83.33&  84.68&  84.00&  80.20&  89.88& 84.76\\
         0.7&  0.3&  \textbf{86.92}&  \textbf{97.50}&  86.53&  88.06&  87.29&  82.76&  91.51& 86.92\\
         0.8&  0.2&  85.79&  \textbf{97.50}&  81.86&  83.71&  82.78&  77.34&  84.69& 80.85\\
         0.9&  0.1&  85.42&  \textbf{97.50}&  83.31&  85.32&  84.30&  79.43& 88.04& 83.51\\
         1.0&  0.0&  83.81&  94.29&  85.49&  \textbf{88.39}&  86.92&  81.36&  89.98& 85.46\\\hline
    \end{tabular}}
    \label{table:outputratio}
\end{table*}

As presented in Table~\ref{table:cuhk} and Figure~\ref{fig:CUHK-MRG}, on the CUHK-MRG dataset, our model exhibits the best performance in all metrics except for the edit score, and increases the average pre-class precision and recall metrics by around 2\% compared to the second-highest scores. When comparing with single-modal methods, we find that, although using only single-modal data might achieve good performance in terms of accuracy (e.g., TCN, TeCNO, TMRNet), it cannot perform well across all metrics. This is because a single modality does contain discriminative information for workflow recognition. However, when both kinematic and vision modalities are used, the model gains more complementary information. Only by using multiple modalities together can we significantly enhance overall recognition performance. When using two modalities, Table~\ref{table:cuhk} presents that all metrics of several multimodal models are at high levels. Compared to single-modality methods, the fusion of multimodal data can explore the information in the dataset more deeply and help the model make accurate recognition. Our GRAD framework achieves the best overall accuracy of 92.83\%, which is 6.42\% higher than the LSTM model using kinematics data.

Besides, we note that some of the edit scores reach the performance of 100\%, and the reason may be that the CUHK-MRG dataset is a small dataset with 12 sequences being collected, and every site follows the same sequential operation. The edit score assesses the model's ability to accurately predict the sequence of action segments, regardless of minor temporal variations. Therefore, when the model is small and trained for an adequate number of epochs, it becomes straightforward for the model to memorize this order. There is a possibility that the model is not aware of the sequential order of the two phrases, but rather subconsciously gives the specific sequence. Consequently, a model's edit score can achieve 100\% on a small dataset.

\subsubsection{Ablation Study on the Proposed Modules}
Table~\ref{table:component} presents the ablation study of the proposed modules. We (i) exclude the calibration module, (ii) remove the visual feature disentanglement, and (iii) remove the vision-kinematic adversarial training. Firstly, we can see that for the baseline in the first row, introducing any new technique results in a performance boost, improving most of the metrics. Besides, when the calibration module is introduced in our model, the performance on average per-class precision, recall, and F1 scores always show significant growth, indicating that the model benefits from the confidence boost provided by calibration, making the network more reliable and enhancing the model’s overall performance. Additionally, adding visual feature disentanglement to different models consistently improves performance, with stable advancements across all metrics. Therefore, the results showcase that using feature disentanglement effectively captures extensive information from the video, leading to a better understanding of the workflow process and improved accuracy. Furthermore, when vision-kinematic adversarial training is added with baseline and visual feature disentanglement, most metrics show a slight increase. This suggests that by projecting vision and kinematic features into joint embedding space, adversarial training can narrow down the modality gap between two inputs and eventually boost performance. Overall, Table~\ref{table:component} demonstrates the effectiveness of our proposed submodules, achieving the best performance only when combined. Additionally, Table~\ref{table:left_right} verifies whether we should model the kinematics of the left and right hands together or separately. The results showcase that when we model the left and right kinematic information together and construct it as a joint graph node, the model shall achieve the best performance.

\begin{table*}
    \centering
    \caption{Ablation study of the source and target modality in adversarial training on the MISAW dataset. `K' denotes kinematic modality. `V', `W', and `F' denote the $x_t^i$, $x_t^w$, and $x_t^f$ from the visual modality, respectively.}
    \setlength{\tabcolsep}{12pt} 
    \resizebox{\textwidth}{!}{
    \begin{tabular}{c|c|cccccccc}
        \hline
        Source Modality& Target Modality  &  Acc &  Edit &  OP &  OR &  OF1 &  CP &  CR & CF1 \\ \hline
         K&   V&  82.97&  96.25&  85.83&  86.23&  86.03&  83.48&  89.15& 83.78\\
         K &  V+W+F&  84.07&  94.14&  84.08&  85.43&  84.75&  82.48&  87.92& 82.72\\
         V&  K&  82.68&  86.17&  86.03&  \textbf{88.39}&  87.18&  \textbf{85.00}&  89.46& 85.93\\
         V+W+F &   K& \textbf{86.87}&  \textbf{97.50}&  \textbf{86.83}&  88.23&  \textbf{87.52}&  83.07&  \textbf{91.66}& \textbf{87.16}\\\hline
    \end{tabular}}
    \label{table:ad_modality}
\end{table*}

\begin{table*}
    \centering
    \caption{Ablation study of the loss function ratio on the MISAW dataset, e.g., the $\gamma$ and $\delta$ in the Equation~(\ref{equ:loss}).}
    \setlength{\tabcolsep}{12pt} 
    \resizebox{\textwidth}{!}{
    \begin{tabular}{c|c|cccccccc}
        \hline
        $\mathcal{L}_{CCE}$ $(\gamma)$ & $ \mathcal{L}_{AL}$ $(\delta) $ &  Acc &  Edit &  OP &  OR &  OF1 &  CP &  CR & CF1 \\ \hline
         1&  0&  85.41&  \textbf{97.50}&  83.02&  85.16&  84.08&  79.61&  89.30& 84.17\\
         0.95&  0.05&  83.62&  94.17&  85.96&  87.90&  86.92&  82.13&  91.00& 86.34\\
         0.9&  0.1&  \textbf{86.87}&  \textbf{97.50}&  \textbf{86.83}&  \textbf{88.23}&  \textbf{87.52}&  \textbf{83.07}&  \textbf{91.66}& \textbf{87.16}\\
         0.85&   0.15& 84.68&  \textbf{97.50}&  83.07&  83.87&  83.47&  79.31&  86.83& 82.90\\
         0.8&  0.2&  83.93&  \textbf{97.50}&  83.30&  84.50&  83.90&  81.17&  86.33 & 83.67\\\hline
    \end{tabular}}
    \label{table:loss}
\end{table*}

\begin{table*}
    \centering
    \caption{Ablation study of the $\lambda$ in the Equation~(\ref{equ:ca_loss}) of calibration loss function on the MISAW dataset.}
    \setlength{\tabcolsep}{15pt} 
    \resizebox{\textwidth}{!}{
    \begin{tabular}{c|cccccccc}
        \hline
        $\lambda$ &  Acc &  Edit &  OP &  OR &  OF1 &  CP &  CR & CF1 \\ \hline
         0.01&  84.12&  97.14&  80.79&  82.74&  81.75&  81.42&  87.76& 84.47\\
         0.02&  \textbf{86.87}&  \textbf{97.50}&  \textbf{86.83}&  \textbf{88.23}&  \textbf{87.52}&  83.07&  \textbf{91.66}& \textbf{87.16}\\
         0.03&   84.97&  95.56&  83.25&  84.19&  83.72& 83.37&  89.82& 86.47\\
         0.04&    84.93&  97.14&  83.65&  85.81&  84.71&  \textbf{83.55}&  89.32& 86.34\\
         0.05&    86.10&  \textbf{97.50}&  83.31&  85.32&  84.44&  83.31&  89.02&  86.07\\\hline
    \end{tabular}}
    \label{table:ca_loss}
\end{table*}

\subsubsection{Ablation Study on Visual Disentanglement}
To understand how visual feature disentanglement improves the model, we conduct an ablation analysis on different feature disentanglement domains. The experimental results are shown in Table~\ref{table:des}. The results indicate that introducing the wavelet transform alone significantly boosts performance, while using the Fourier amplitude spectrum alone does not improve it. Wavelet transform better identifies multi-scale local features in visual signals and better understands edge textures, enhancing the model's ability to recognize details. In contrast, the Fourier amplitude spectrum emphasizes global structural information in visual signals. Using Fourier alone may lead to some performance loss, but combining the local multi-scale understanding from the wavelet domain with the global structure understanding from Fourier provides the model with more diverse information. This complementary use results in notable performance improvement. In summary, the spatial, wavelet, and Fourier domains we use are essential for the model's effectiveness, as they contribute to information extraction, feature utilization, and an overall understanding of our model's workflow.

\subsubsection{Ablation Study on the Graph Network}
\label{exper:graph_network}
To justify the superiority of using GAT in our graph learning part, we perform ablation experiments over different graph networks on the MISAW dataset, with GCN and RGCN as comparison networks.
In GCN, the input adjacent matrix is given by relating kinematic, spatial, wavelet, and Fourier modalities together as one initial setting for graph representation learning. 
RGCN, on the other hand, can define more adjacent matrices which will broaden the limited correlation within GCN. In this case, we can construct four different relationships based on our features. For example, the kinematic features are influenced by spatial, wavelet, and Fourier features. The setting is extended to the other three features in the same manner. 
As for GAT, the utilization of the attention mechanism enables the network to assign varying weights to all related nodes according to their respective levels of influence. Table~\ref{table:gcn} shows that by assigning varying levels of importance to nodes dynamically, GAT can learn the most valuable feature representations. However, when a fixed relationship is assigned to the graph network, certain inherent weight relationships may remain unseen, leading to suboptimal performance of GCN and RGCN.

\subsubsection{Ablation Study on the Ratio of Vision-Kinematic Feature \& Graph Embeddings}
\label{exper:abl_embedding_fusion}
For the weighted fusion of vision-kinematic feature embeddings and graph output embeddings in Equation~(\ref{equ:feature_output}), we further investigate how the parameters $\alpha$ and $\beta$ in Equation~(\ref{equ:feature_output}) achieve a perfect balance. Specifically, we integrate the feature embeddings from visual and kinematic data before the graph network with the embedding output by the graph network based on Equation~(\ref{equ:feature_output}). The final input to the workflow prediction decoder is the weighted sum of these two different outputs. Our objective is to seek the best performance of our model by finding the optimal balance between vision-kinematic feature output and graph output through ablation experiments.
Through a thorough comparative analysis of various ratios in Table~\ref{table:outputratio}, we find that the ratio of 0.7 for $\alpha$ and 0.3 for $\beta$ results in the highest accuracy, but it slightly underperforms in all other evaluation metrics compared to the ratio of 0.3 for $\alpha$ and 0.7 for $\beta$. Different ratio configurations exhibit respective advantages on different indicators, demonstrating that both the feature and graph output embeddings contribute specifically to model performance. Finally, $\alpha = 0.3$ and $\beta = 0.7$ achieve the highest performance across multiple metrics; therefore, we use this configuration as the final parameters for our model.

\begin{table*}
    \centering
    \caption{Ablation study on different fusion strategies compared with our graph learning strategy. All experiments employ the same backbone, feature disentanglement, adversarial learning, and loss function. "Attn": Attention.}
    \setlength{\tabcolsep}{12pt} 
    \resizebox{\textwidth}{!}{
    \begin{tabular}{c|cccccccc}\hline
        Methods &  Acc &  Edit &  OP &  OR &  OF1 &  CP&  CR& CF1 \\ \hline
        Add &  84.60&  94.55&  81.09&  83.71&  82.38&  75.88&  83.23& 79.39\\ 
        Concat&  83.73&  90.83&  81.07&  82.90&  81.98&  76.63&  83.60& 79.96\\ 
        Cross Attn (V2K)&  70.61&  81.81&  83.62&  87.26&  85.40&  80.64&  82.23& 81.43\\ 
        Cross Attn (K2V)&  71.99&  81.57& \textbf{88.68}&  84.68&  86.63&  85.37&  88.05& 86.69\\ 
        Self Attn&  76.44&  86.33&  85.93&  82.74&  84.31&  82.24&  86.23& 84.19\\ 
        Gated Fusion&  84.24&  95.00&  77.54&  81.29&  79.37&  66.33&  72.91& 69.46\\ 
        BAN&  74.66&  89.76&  86.54&  84.03&  85.27&  83.53&  88.76& 86.07\\ 
        TDA&  74.78&  88.17&  88.43&  86.29&  87.35&  \textbf{84.46}&  88.80& 86.57\\ 
        AFF&  85.82&  95.00&  86.42&  87.26&  86.84&  74.56&  82.53& 78.34\\ 
        iAFF&  83.05&  96.25&  85.65&  87.58&  86.60&  80.98&  87.46& 84.10\\ \hline
        GRAD (\textbf{Ours}) &  \textbf{86.87}&  \textbf{97.50}&  86.83&  \textbf{88.23}&  \textbf{87.52}&  83.07&  \textbf{91.66}& \textbf{87.16}\\ \hline
    \end{tabular}}
    \label{table:fusion}
\end{table*}

\begin{figure*}[h]
    \centering
    \includegraphics[width=1.9\columnwidth]{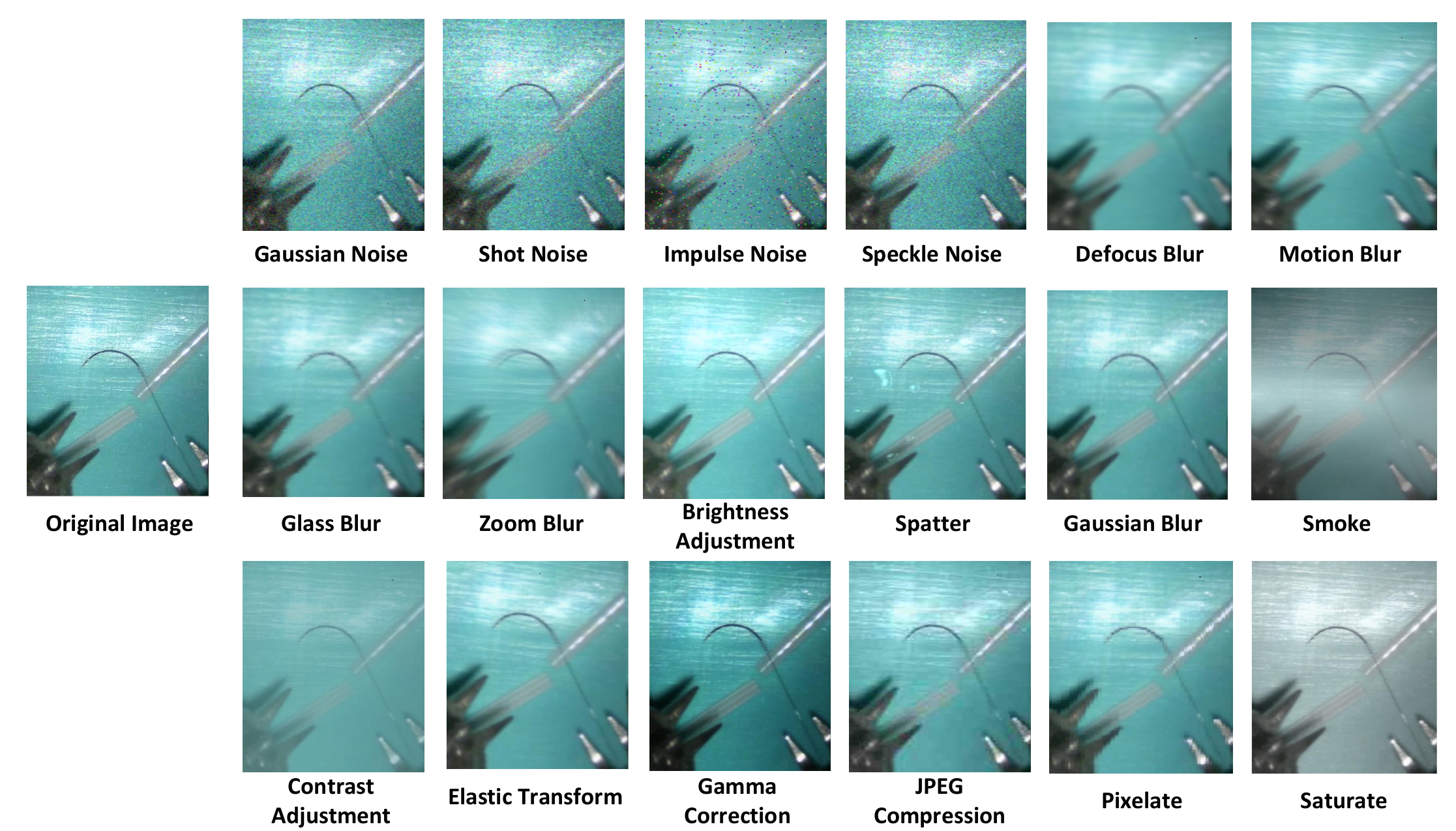}
    \caption{An example of corrupted data visualization for robustness assessment, featuring four different types ranging from noise to digital damage.}
    \label{fig:robust_visual}
\end{figure*}

\begin{figure*}
    \centering
    \includegraphics[width=0.8\linewidth]{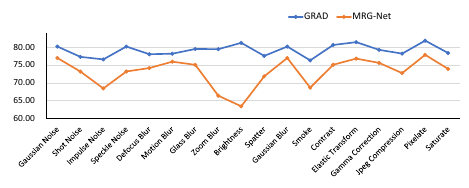}
    \caption{Robustness experiments on the MISAW dataset under 18 different corruptions.}
    \label{robust_fig2}
\end{figure*}

\begin{figure}
    \centering
    \includegraphics[width=\linewidth]{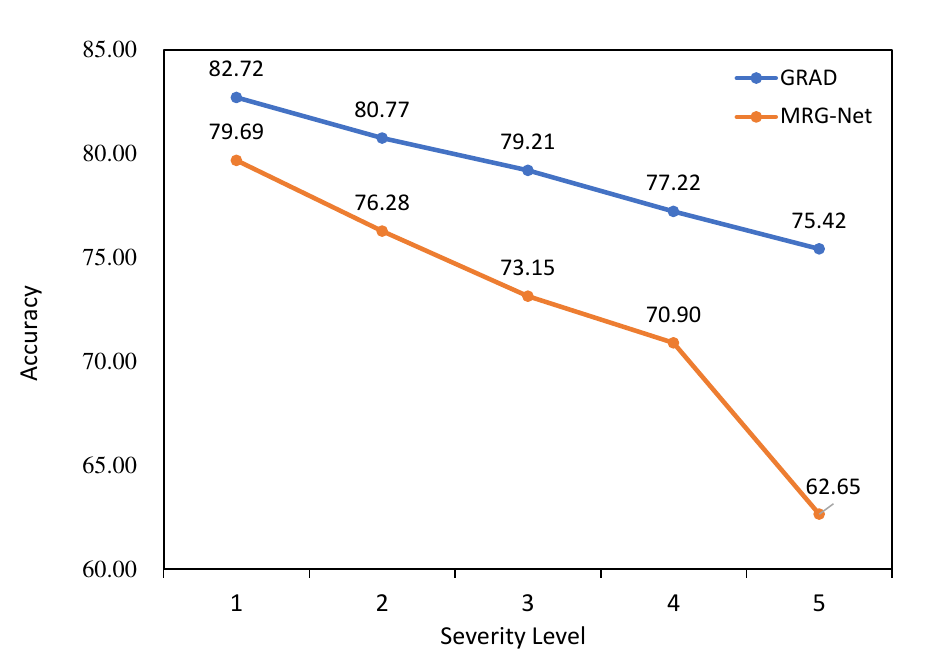}
    \caption{Robustness experiments on the MISAW dataset at each severity level.}
    \label{robust_fig1}
\end{figure}

\begin{table*}[t]
    \caption{Robustness comparison experiments of our GRAD with MRG-Net across 18 different types of pollution and 5 levels of severity on the MISAW dataset.}
    \centering
    \resizebox{\textwidth}{!}{
    \begin{tabular}{ c | c | cc | cc | cc | cc | cc } \hline
    \multicolumn{2}{c|}{Severity} & \multicolumn{2}{c|}{Level 1} & \multicolumn{2}{c|}{Level 2}  & \multicolumn{2}{c|}{Level 3}  & \multicolumn{2}{c|}{Level 4}  & \multicolumn{2}{c}{Level 5}  \\ \hline
    \multicolumn{2}{c|}{Model} & MRG~\cite{long2021relational}  & GRAD & MRG~\cite{long2021relational}  & GRAD & MRG~\cite{long2021relational}  & GRAD & MRG~\cite{long2021relational}  & GRAD & MRG~\cite{long2021relational}  & GRAD  \\ \hline
    \multirow{4}{*}{Noise}     & Gauss  & 81.90 & \textbf{83.03} & 79.35 & \textbf{82.58} & 77.87 & \textbf{80.65} & 75.15 & \textbf{80.19} & 70.93 & \textbf{74.96}  \\ \cline{2-12}
    & Shot & \textbf{80.73} & 80.24 & 76.80  & \textbf{77.44}  & 71.45 & \textbf{77.19} & 68.17 & \textbf{75.76} & 68.86 & \textbf{76.15} \\ \cline{2-12}
    & Impulse & 76.72 & \textbf{81.02} & 67.87 & \textbf{77.56} & 66.28 & \textbf{74.86} & 67.28 & \textbf{74.74} & 64.07 & \textbf{75.00}  \\ \cline{2-12} 
    & Speckle & 79.20 & \textbf{83.15} & 75.85 & \textbf{81.57} & 73.15 & \textbf{81.02} & 70.76 & \textbf{79.71} & 67.20 & \textbf{75.91}  \\ \hline
    \multirow{5}{*}{Blur} & Defocus & 77.19 & \textbf{80.71} & 76.27 & \textbf{79.09} & 72.35 & \textbf{77.77} & 72.17 & \textbf{76.56} & 73.06 & \textbf{76.42}  \\ \cline{2-12}
    & Motion & 79.34 & \textbf{82.35} & 76.70 & \textbf{79.95} & 75.66 & \textbf{77.41} & 75.36 & \textbf{75.53} & 72.93 & \textbf{76.06}  \\ \cline{2-12}
    & Glass & 80.52 & \textbf{83.25} & 77.68 & \textbf{82.55} & 75.04 & \textbf{79.49} & 72.48 & \textbf{77.26} & 69.87 & \textbf{75.42}  \\ \cline{2-12}
    & Gauss & 81.90 & \textbf{83.03} & 79.35 & \textbf{82.58} & 77.87 & \textbf{80.65} & 75.15 & \textbf{80.19} & 70.93 & \textbf{74.96}   \\ \cline{2-12}
    & Zoom & 74.58 & \textbf{81.71} & 71.36 & \textbf{80.72} & 64.11 & \textbf{79.62} & 62.50 & \textbf{78.62} & 59.39 & \textbf{76.99}  \\ \hline
    \multirow{3}{*}{Occlusion} & Brightness & 81.76 & \textbf{85.30} & 80.97 & \textbf{83.17} & 78.73 & \textbf{80.54} & 74.84 & \textbf{78.13} & 69.70 & \textbf{79.44}   \\ \cline{2-12}
    & Spatter & 77.95 & \textbf{82.31} & 75.29 & \textbf{79.01} & 74.71 & \textbf{79.14} & 71.18 & \textbf{76.88} & 63.05 & \textbf{70.83}   \\ \cline{2-12}
    & Smoke & 81.84 & \textbf{82.33} &  74.49 & \textbf{78.29} & 66.27 & \textbf{76.14} & 62.13 & \textbf{73.32} & 58.52 & \textbf{71.77}   \\ \hline
    & Contrast & 80.73 & \textbf{84.79} & 78.80 & \textbf{83.61} & 76.84 & \textbf{81.42} & 72.18 & \textbf{77.04} & 67.18 & \textbf{76.76}   \\ \cline{2-12}
    \multirow{4}{*}{Digital}& Elastic & 80.68 & \textbf{83.63} & 79.75 & \textbf{82.82} & 77.79 & \textbf{82.02} & 74.44 & \textbf{80.53} & 71.68 & \textbf{78.56}   \\ \cline{2-12}
    & Gamma & 82.46 & \textbf{84.19} & 76.69 & \textbf{82.37} & 75.37 & \textbf{81.66} & 73.99 & \textbf{76.87} & 69.72 & \textbf{71.65}   \\ \cline{2-12}
    & Jpeg & 78.84 & \textbf{81.38} & 78.03 & \textbf{79.28} & 75.09 & \textbf{75.58} & 68.68 & \textbf{73.23} & 63.15 & \textbf{81.90}   \\ \cline{2-12}
    & Pixelate & 81.12 & \textbf{83.59} & 80.97 & \textbf{83.11} & 79.15 & \textbf{82.64} & 75.02 & \textbf{81.18} & 73.44 & \textbf{79.22}   \\ \cline{2-12}
    & Saturate & 81.04 & \textbf{83.42} & 77.29 & \textbf{83.04} & 75.85 & \textbf{81.21} & 70.86 & \textbf{73.27} & 64.74 & \textbf{71.30}   \\ \hline
    \multicolumn{2}{c|}{Mean} & 79.95 & \textbf{82.77} & 76.67 & \textbf{81.00} & 73.81 & \textbf{79.40} & 70.60 & \textbf{88.15} & 63.12 & \textbf{75.92}   \\ \hline
    \end{tabular}
    \label{table:robust}}
\end{table*}

\subsubsection{Ablation Study on the Source and Target Modalities in Adversarial Training}
\label{exper:abl_target}
To investigate how treating different inputs as source and target modalities affects the model's performance, we conduct related experiments in Table~\ref{table:ad_modality}. By designating various modalities as source and target, adversarial learning enables the projection of distinct modalities into a shared embedding space, aligning target distributions with source distributions within a modality-invariant space. The results in Table~\ref{table:ad_modality} indicate that when kinematic features are used as the source modality and visual representations as the target modality, the model exhibits inferior performance across most metrics compared to other configurations. This is likely due to the inherent difficulty of mapping visual distributions onto kinematic distributions, as visual representations contain more complex information. Additionally, the wavelet and Fourier representations, which share similarities with spatial representations, allow the model to process them more effectively in combination. Overall, the results demonstrate that treating visual representations from three domains as the source modality and kinematic data as the target modality enables the model to achieve satisfactory performance.

\subsubsection{Ablation Study of the Loss Function}
\label{exper:abl_loss}
Experiments in Table~\ref{table:loss} analyze the ratio between adversarial loss $\mathcal{L}_{AL}$ and calibrated classification loss $\mathcal{L}_{CCE}$ to find the ideal balance. As shown in Table~\ref{table:loss}, the model performs poorly when the adversarial loss proportion is too high. This suggests that overemphasizing the reduction of the modality gap may cause the model to neglect the workflow recognition task, leading to poor performance. However, when the adversarial loss ratio is too low or absent, the model also performs poorly on most metrics, indicating the need for adversarial loss to help approximate and integrate both modalities. Ultimately, by analyzing the results, we find the best performance with a ratio of 0.9 for $\mathcal{L}_{CCE}$ and 0.1 for adversarial loss $\mathcal{L}_{AL}$. This demonstrates the importance of reducing the gap between different modalities and highlights the contribution adversarial loss makes to our model. Additionally, we explored the setting of the regularization coefficient $\lambda$ in Equation~(\ref{equ:ca_loss}). Table~\ref{table:ca_loss} shows that we achieve the best results when $\lambda = 0.02$.

\subsubsection{Ablation Study on Different Fusion Strategies}
In Table~\ref{table:fusion}, we compare our GRAD method with other multimodal fusion techniques to demonstrate its effectiveness in integrating visual and kinematic data for robotics. We evaluate different fusion methods while keeping other modules consistent. The methods include Matrix Addition, Concatenation, Co-Attention (V2K: vision as attention queries; K2V: kinematics as attention queries)~\cite{lin2022cat}, Self-Attention~\cite{vaswani2017attention}, Gated Fusion~\cite{bai2023surgical}, BAN~\cite{kim2018bilinear}, TDA~\cite{anderson2018bottom}, AFF \& iAFF~\cite{dai2021attentional}. Specifically, we replace our graph-based fusion with these different methods, while retaining the visual feature disentanglement and calibration components to ensure a fair comparison. The results show that attention-based methods like AFF and iAFF achieve decent performance, but simpler combination methods or gated networks yield better outcomes. This indicates that direct attention-based fusion might not be optimal, and further exploration of the interaction between kinematics and vision is needed. Our graph representation approach achieves the best results.

\subsection{Robustness}

To thoroughly evaluate the robustness of our proposed GRAD framework, we conduct extensive experiments under diverse image corruptions using the MISAW dataset, performing a comparative analysis with MRG-Net~\cite{long2021relational}. For a comprehensive experiment, we assess visual data exposed to 18 distinct corruption types, categorized into four main groups: Noise, Blur, Occlusion, and Digital corruptions, from a 2D image corruption benchmark~\cite{hendrycks2019benchmarking}. 
Figure~\ref{fig:robust_visual} provides a visualization of each corruption type. 
Besides, each corruption type is applied across five severity levels, enabling a thorough observation of model performance under escalating corruption intensities.

Figure~\ref{robust_fig1} shows the average results across 18 corruption types at each corruption severity intensity. We can find that the performance of both methods gradually decreases as the severity level of corruption increases. However, our model consistently outperforms the baseline MRG-Net across all severity levels, and exhibits significantly less performance decline. This outcome underscores the superior stability of our model in the presence of perturbations from visual corruption. Through our proposed GRAD framework, the model can better extract detailed information from corrupted images, thereby demonstrating greater robustness. Furthermore, we evaluate the average results across different severity levels, which are provided in Figure~\ref{robust_fig2}. When the severity intensity reaches the highest level, our model achieves an accuracy exceeding 75\% across most categories of image corruption and surpasses 80\% in certain categories. In contrast, MRG-Net performs lower than 75\% in all corruption types. This demonstrates our model's remarkable performance in handling various forms of image degradation. Detailed results for each type of corruption are shown in Table~\ref{table:robust}, where the comparative analysis reveals substantial performance for our GRAD model across various corruption scenarios. Our model consistently attains higher accuracy than MRG-Net, a testament to the efficacy of incorporating multimodal adversarial training, visual disentanglement, and calibration loss. These methods markedly bolster the robustness of our model, enhancing its resilience to image corruption challenges and leading to more reliable real-world application ability.

\subsection{Discussion and Limitations}

Overall, our proposed GRAD framework consistently achieves the best performance compared to existing state-of-the-art methods, even on corrupted data. This demonstrates that our framework excels in performance and is highly robust to handle data corruption. With our feature disentanglement and effective graph representation strategy, our model can efficiently model representations from different modalities and find the optimal multimodal balance. Additionally, with our graph calibration and multimodal adversarial training, our framework can robustly work on low-quality data in different scenarios, demonstrating great potential for real-world applications.

While our proposed model demonstrates improved performance on both datasets, limitations still exist. (i) Firstly, the short duration of specific phrases in the dataset leads to difficulties in accurate recognition during training, causing a bias toward more frequent categories and resulting in potential misclassifications. Future work should address this data imbalance to enhance the model’s ability to learn short-duration phrases. (ii) Secondly, training with multimodal data requires significant time and resources, including precise annotation methods for aligning visual signals, kinematic data, and labels, as well as additional computational costs for training and inference. Evaluating the trade-off between performance gains and resource consumption is essential, alongside exploring more efficient strategies at both the data and algorithm levels. (iii) Lastly, all experiments in this study are conducted on benchmark datasets rather than real surgical scenes, due to the lack of a comprehensive multimodal dataset for surgical applications. This limits the evaluation of our proposed framework under real-world clinical conditions, where unique challenges such as lighting, occlusions, and sensor variability exist. Future efforts should focus on creating high-quality multimodal datasets tailored to surgical scenarios to validate and improve the model's robustness and applicability.

\section{Conclusion}
In this paper, we present a robust workflow recognition model, GRAD, offering a comprehensive and robust solution for surgical workflow recognition by effectively integrating multiple data modalities. Our framework integrates visual feature disentanglement across different domains with graph-based representation learning, effectively capturing complementary information between visual and kinematic modalities. Furthermore, we propose the Vision-Kinematic Adversarial strategy to align multimodal feature distributions and the contextual calibrated prediction decoder to enhance robustness and confidence in modality representations, even under data corruption or domain shifts.
The proposed GRAD framework outperforms existing single-modality and multi-modality methods in surgical workflow recognition tasks on two public datasets. Extensive ablation experiments further demonstrated the effectiveness of the individual components. The disentanglement of spatial and frequency features in visual data and the temporal modeling of kinematic data significantly enhanced workflow recognition accuracy. Moreover, the GRAD model exhibits high robustness against data corruption, showing slower performance degradation in robustness tests. Overall, our GRAD framework presents strong reliability and adaptability by capturing intricate patterns and relationships inherent in visual and kinematic data, proving its potential for real-world clinical applications.


\bibliographystyle{cas-model2-names}

\bibliography{cas-refs}


\end{document}